\documentclass{article}
\usepackage[table, dvipsnames]{xcolor}
\usepackage[utf8]{inputenc}
\usepackage{authblk}
\usepackage{setspace}
\usepackage[margin=1in]{geometry}
\usepackage{graphicx}
\graphicspath{ {./figures/} }
\usepackage{subcaption}
\usepackage{amsmath}
\usepackage{booktabs}
\usepackage{arydshln} 
\usepackage{multirow}
\usepackage{mathtools}  
\usepackage{xfrac}
\usepackage{wrapfig}

\usepackage[style=nejm, 
citestyle=numeric-comp,
sorting=none]{biblatex}
\title{Adversarial Attacks on Foundational Vision Models}

\author{Nathan Inkawhich}
\author{Gwendolyn McDonald}
\author{Ryan Luley}

\affil{Air Force Research Laboratory}

\date{July 10, 2023}

\begin{document}

\maketitle

\begin{abstract}
Rapid progress is being made in developing large, pretrained, task-agnostic foundational vision models such as CLIP, ALIGN, DINOv2, etc.
In fact, we are approaching the point where these models do not have to be finetuned downstream, and can simply be used in zero-shot or with a lightweight probing head.
Critically, given the complexity of working at this scale, there is a bottleneck where relatively few organizations in the world are executing the training then sharing the models on centralized platforms such as HuggingFace and \texttt{torch.hub}.
The goal of this work is to identify several key adversarial vulnerabilities of these models in an effort to make future designs more robust.
Intuitively, our attacks manipulate deep feature representations to fool an out-of-distribution (OOD) detector which will be required when using these open-world-aware models to solve closed-set downstream tasks. 
Our methods reliably make in-distribution (ID) images (w.r.t.~a downstream task) be predicted as OOD and vice versa while existing in extremely low-knowledge-assumption threat models.
We show our attacks to be potent in whitebox and blackbox settings, as well as when transferred \textit{across} foundational model types (e.g., attack DINOv2 with CLIP)!
This work is only just the beginning of a long journey towards adversarially robust foundational vision models.
\end{abstract}


\section{Introduction} \label{sec:introduction}




Next-generation machine learning (ML) models in vision are right at our doorstep.
Models such as CLIP \cite{radford21a}, SWAG \cite{singh2022revisiting}, DINOv2 \cite{oquab2023dinov2}, etc., offer unparalleled flexibility in standard and challenging operating conditions (e.g., low-shot, distribution shifts) and the workflows they allow for make ``supervised learning from scratch'' look old-fashioned (e.g., zero-shot).
Powering this trend are the big-AI houses (e.g., Meta, Google, OpenAI, etc.) which are testing scaling laws by training large models on web-scale datasets using compute clusters with 100s of GPUs (it's even become relevant to document the environmental impacts of training \cite{oquab2023dinov2}).
Generally, the recipe being used is more data + more compute + better distributed training methods (e.g., optimizers) + algorithmic advances in un/self/semi-supervised learning (to leverage \textit{unlabeled} data) + transformers (although some may debate this one \cite{tian2023designing}).
With these algorithms, at this scale, cool things happen: emergent functions like zero-shot learning \cite{radford21a} and open-vocabulary recognition/detection \cite{abs-2306-09683}; models learn task-agnostic representations which are highly robust, transferable, and semantically meaningful; and multi-modal learning is somewhat straightforward \cite{radford21a, imagebind}.
The new workflow for ``solving'' a downstream task will soon be to download the best foundation model your hardware can support then use a very small amount of task-specific information to partition the model's feature space, rather than learning a whole feature extractor and classifier from scratch.

\begin{wrapfigure}{r}{.5\linewidth}
    \vspace{-4mm}
    \centering
    \includegraphics[width=0.99\linewidth]{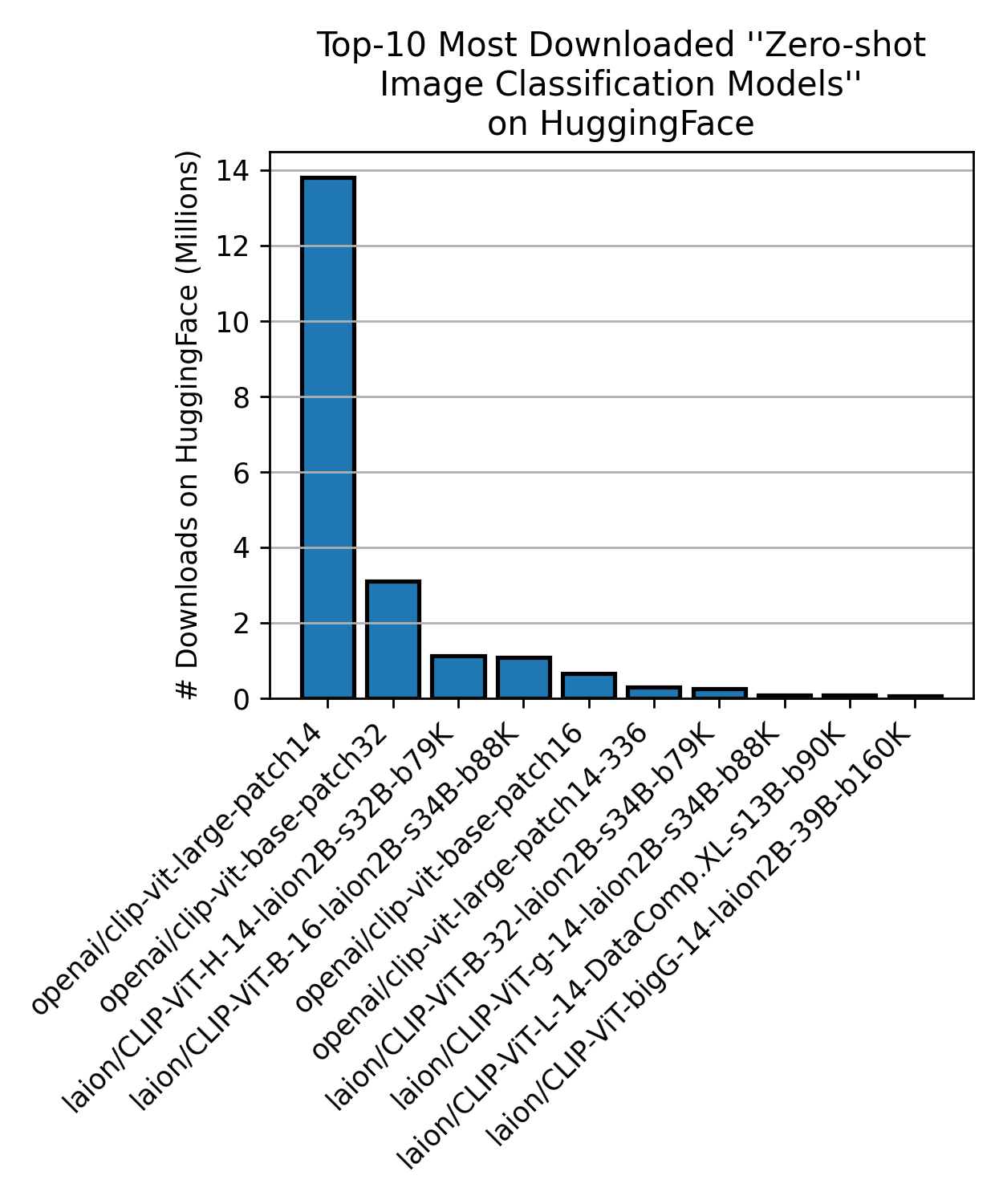}
    \caption{Download stats for the top-10 most popular ``Zero-shot Image Classification'' models on HuggingFace as of June 9, 2023.}
    \label{fig:hf_downloads}
\end{wrapfigure}

As these models and methods improve it will be tempting to use them off-the-shelf \textbf{without} customization. 
We're already seeing this workflow being promoted as desirable: ``a strong property of our approach [DINOv2] is that finetuning is optional'' \cite{oquab2023dinov2}.
This trend directly motivates our work because we believe that it exposes a severe adversarial vulnerability to inference time attacks, where a clever adversary can actually near a whitebox attack threat model.
Given the centralized distribution of these models on HuggingFace and \texttt{torch.hub} and low diversity of organizations capable training at this scale, it's feasible that an adversary can make an accurate guess at the exact model their target is using. 
Figure~\ref{fig:hf_downloads} displays the HuggingFace download statistics as of June 9, 2023 for the top-10 most popular ``Zero-shot Image Classification'' models. 
Obviously, a few of the models are significantly more popular than the rest, which is the exact information an adversary can exploit.
With straightforward techniques like ensembling, the attacker can further mitigate their risk of picking the wrong model.
Finally, note the lack of diversity in model architectures.
The two main entities, Laion and OpenAI, are both are using standard ViT variants, which is a another design trend that can be exploited and likely makes blackbox transfer attacks more potent in the event the adversary does not guess correctly.

Our goal in this work is to raise awareness of this budding adversarial vulnerability by showing how simple adversarial attacks under very low-knowledge threat models can degrade the performance of s.o.t.a. foundational models downstream.
One of the only key assumptions we make is that the foundational model is not being finetuned, and instead is being used in zero-shot or with some supervisedly-learned classifier head (e.g., linear probing).
Our attack intuition is to target the out-of-distribution (OOD) detector component of these systems, which will almost surely be required as these open-world-aware foundational models are utilized to solve closed-set downstream tasks (more discussion on this in Section~\ref{sec:methods}).
OOD detection has been an extremely active area of research lately and many methods are easily adoptable in these settings \cite{hendrycks2017a, ming2022delving, zhang2023openood}.
Functionally, our attack perturbs images to manipulate their feature space representations.
We develop an ID$\rightarrow$OOD attack to perturb an in-distribution (ID) input (w.r.t. the downstream task) s.t. it will get flagged and rejected by the OOD detector, causing a silent false negative.
We also develop an OOD$\rightarrow$ID attack, which perturbs an OOD input (again, w.r.t. the downstream task) s.t. it will get predicted with high confidence as one of the task's ID classes, causing a silent false positive.

\begin{figure}[t]
    \centering
    \includegraphics[width=0.7\textwidth]{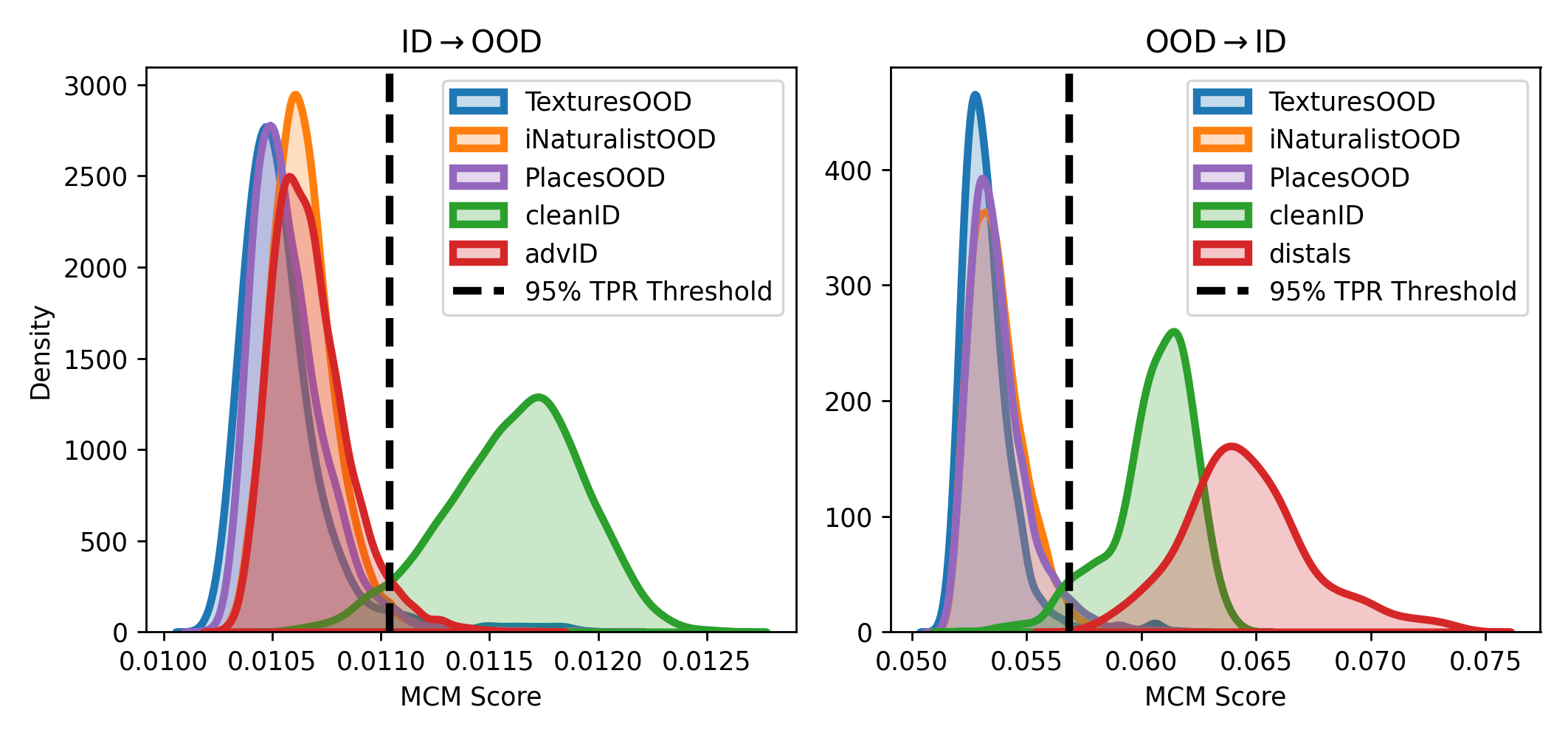}
    \caption{Intuition for how our attacks manipulate the OOD detection scores of models performing downstream tasks (left subplot is ID$\rightarrow$OOD, right subplot is OOD$\rightarrow$ID, and models are being used in zero-shot with an MCM detector \cite{ming2022delving}).}
    \label{fig:attack_intuition}
\end{figure}

Figure~\ref{fig:attack_intuition} shows conceptually how our attacks manipulate the recognition system.
In both subplots, a CLIP model \cite{radford21a} is being used in zero-shot to solve a downstream ID task with an MCM OOD detector \cite{ming2022delving} predicting ID vs OOD scores for each input (x-axis).
Higher MCM scores indicate higher ID-ness and lower scores mean higher OOD-ness. 
From the plots, this detector is doing a good job in the usual evaluation settings, as the true ID ({\color{ForestGreen}cleanID}) data is consistently predicted with higher scores than the naturally occurring OOD data {\color{blue}TexturesOOD}, {\color{orange}iNaturalistOOD} and {\color{violet}PlacesOOD} (i.e., AUROC~$>95\%$ and FPR95~$<10\%$ in OOD detection verbiage).
The vertical dashed line indicates the MCM score threshold that yields a 95\% True Positive Rate (TPR) which is commonly considered in the literature.
The left subplot shows the impact of an ID$\rightarrow$OOD attack where {\color{ForestGreen}cleanID} inputs are perturbed to create the {\color{red}advID} distribution. 
Nearly all {\color{red}advID} samples fall below the 95\% TPR threshold and are indistinguishable from the naturally occurring OOD data in terms of MCM score, marking a successful attack. 
The right subplot shows the impact of an OOD$\rightarrow$ID attack. 
In this work, we use random noise images as the starting point of the optimization which we call distal adversaries \cite{inkawhich_neurips20, Stutz0S20} (see Figure~\ref{fig:distal_samples} for visual samples). 
This attack is also very powerful as the {\color{red}distals} almost always get predicted with scores above the 95\% TPR threshold.


Overall, our contributions in this work are as follows:
\begin{itemize}
    \item We observe several key trends in today's foundational model research that expose serious adversarial vulnerabilities requiring very little knowledge on the part of the adversary;
    \item We develop ID$\rightarrow$OOD and OOD$\rightarrow$ID adversarial attacks and show both to be powerful in whitebox and blackbox transfer attack settings;
    \item We show that our attacks can reliably transfer across foundational model training algorithms (from CLIP to DINOv2 and SWAG), across backbone architectures (from ViT's to ConvNeXT's), and across different classifier-head schemes (zero-shot, linear probe, and kNN);
    \item We provide several concrete suggestions for future work in an attempt to mitigate our found adversarial vulnerability in the next generation of foundational algorithms.
\end{itemize}
An ideal outcome of this work is to get foundational model developers to take the adversarial robustness of their models seriously and to make adopters of current generation models aware of simple ways in which their models can be fooled.

\section{Related Work}

\subsection{Rise of Foundational (Vision) Models} 

At this point in time, it is hard to define exactly what a ``Foundational Model'' in computer vision is.
In our view a key attribute of foundational models is task-agnosticism, meaning the model is not trained to only accomplish a specific task, in a specific set of operating conditions, for a specific application (thus, we do not believe a ResNet trained with supervised learning on ImageNet-1K is foundational). 
Rather, such a model is trained to learn generally good representations of the data which are highly transferable to many potential tasks that may be encountered downstream.

Recently, the rise of Self-Supervised Learning (SSL)~\cite{ssl_cookbook} in computer vision has started to produce task-agnostic models with an odor of foundational-ism.
For example, models like SimCLR~\cite{ChenK0H20}, BYOL~\cite{GrillSATRBDPGAP20}, DINO~\cite{CaronTMJMBJ21}, VICReg~\cite{BardesPL22}, DINOv2~\cite{oquab2023dinov2}, etc., learn representations from unlabeled images and have all shown an impressive amount of flexibility in solving diverse downstream tasks using very little labeled data.
Similarly, models like CLIP~\cite{radford21a}, SWAG~\cite{singh2022revisiting}, and ALIGN~\cite{JiaYXCPPLSLD21} include a vision-language component in the pretraining which facilitates even more downstream flexibility with the introduction of ``zero-shot'' learning (see Figure~\ref{fig:zeroshot_vs_linprobe_overview} in Appendix).
Progress in this area is happening daily, and by the time you are reading this several more models are likely available which build upon these.
Finally, we want to make it clear that in this work we are talking about foundational models for visual ``classification'' tasks, and not models like Segment-Anything \cite{kirillov2023segany} or OWLv2~\cite{abs-2306-09683} which include localization components.

\subsection{Adversarial attacks on foundation/CLIP models}

There is a significant amount of literature discussing ways in which classifiers trained with supervised learning can be adversarially attacked \cite{SzegedyZSBEGF13, GoodfellowSS14, Carlini017, inkawhich_cvpr19, inkawhich_iclr20, inkawhich_neurips20, InkawhichLZYLC21}.
While several important threat models have been identified (e.g., white/gray/black-box) a commonality across most of these methods is that the target model is assumed to be trained to accomplish a specific task.
Thus, the attack methods tend to manipulate the output classification layer of these models.
Due to the task-agnostic nature of foundational models, many of these attacks may have limited applicability because there is no output classification layer to manipulate.
In this work, our attacker's threat model does NOT include knowledge of the entire target model's label space, and thus our attack is unique from these.


Several recent works have discussed attacks on foundational models, including CLIPs.
S.~Fort~\cite{sfort_clipattack1} verified that ``standard'' adversarial attacks work well on CLIP models once they have been adapted for a specific downstream task.
In separate works, S.~Fort~\cite{sfort_clipattack2} and D.~Noever and S.~Noever~\cite{noever_clipattack} consider attacks on multi-modal CLIP neurons by placing text stickers on images such that the CLIP model ``reads'' the text and ignores everything else in the image.
These attacks are operating under a different threat model, as the adversarial manipulations are very obvious.
Uniquely, N.~Carlini~\textit{et al.}~\cite{carlini2023poisoning} recently showed that poisioning attacks on web-scale datasets are possible and can significantly impact the training process of foundational models like CLIP.

There are two works that are notably closest to ours.
First, Y.~Ban and Y.~Dong~\cite{BanD22} design ``pre-trained adversarial perturbations'' for a threat model where target entities download models (e.g., SimCLR, SimCLRv2, MAE, CLIP) from public sites then finetune them to accomplish specific downstream tasks.
Their threat model differs from ours in a few important ways: (1) they assume the target model will be finetuned; (2) they assume the attacker knows exactly which model the target has downloaded to begin finetuning; (3) they strive to design universal perturbations whereas ours are instance-specific; and (4) they assume the attacker has access to a relatively large dataset (which happens to be a sub-set of the pretraining dataset) to help generate their universal perturbations.
We consider our work highly complementary to theirs, as we focus on the setting where the target's foundational model will not be finetuned and where the attacker does not necessarily have to know which model the target has downloaded nor any of the pretraining data.
Second, J.~Zhang~\textit{et al.}~\cite{ZhangYS22} develop the ``collaborative multimodal adversarial attack'' (co-attack) which attacks the image and text modalities of a vision-language model simultaneously.
Using similar intuition to ours, their attacks manipulate the embedding representations of these models, however, they focus on distinct downstream tasks like retrieval, entailment, and grounding and optimize a slightly different signal involving KL-divergence.
Also, their adversaries are always assumed to have whitebox access to the target model (i.e., can directly compute gradients) and their expectation that the adversary can manipulate both text and visual inputs is beyond the assumptions we make.
We also consider our work complementary to theirs in that we both confirm the intuition of moving around in feature space is a productive attack signal. 
However, our work shows the impacts of this manipulation for different downstream tasks, under different assumption sets (like blackbox), and when transferred across different pretraining algorithms.


Worth noting, there are several recent works discussing adversarial defenses for CLIP models. 
C. Mao~\textit{et al.}~\cite{mao_iclr23} and X. Li~\textit{et al.}~\cite{li2023languagedriven} both develop adversarial training schemes which attempt to bolster zero-shot adversarial robustness.
These works consider rather vanilla instantiations of adversaries at small $\epsilon$'s, do not extend straightforwardly to models like DINOv2 which do not have language encoders, and both incur significant losses in clean performance while requiring non-trivial compute to pull off at moderate scales.
In our opinion, unless these techniques are adopted by major model providers, we do not believe an average user is likely to adopt them.

\subsection{OOD Detection and Attacks Against Detectors}

OOD detection has garnered a significant amount of attention lately \cite{hendrycks2017a, liang2018enhancing, 9695222, HendrycksBMZKMS22, ming2022delving, zhang2023openood}.
There are two primary goals in standard OOD detection literature: (1) maintain high classification accuracy on the ID data; and (2) reliably identify anomalous inputs which do not belong to one of the ID categories for the task.
From there, many sub-types of OOD detectors have been developed that work under different assumption sets (e.g., whether or not any OOD samples are available during training time, or, whether or not the practitioner can influence the model's training at all).
For reasons discussed in Section~\ref{sec:methods}, we believe that OOD detection will be a critical function when using foundation models due to the open-world nature of pretraining and the closed-set nature of downstream recognition tasks.

The most relevant sub-type of OOD detector to our work are so called post-hoc detectors, which work with fixed pretrained models.
A very common/powerful standard baseline is a flavor of ``Maximum Softmax Probability'' (MSP) where the OOD score is simply related to the maximum softmax score for the prediction \cite{hendrycks2017a, liang2018enhancing, HendrycksBMZKMS22}.
Another relevant detector to this work is the ``Maximum Concept Matching'' (MCM) \cite{ming2022delving} method, which can be thought of as a variant of MSP when using CLIP models in zero-shot.
In this work we use MSP and MCM OOD detectors in our evaluations, which we believe are competitive with any other post-hoc methods practitioners may use \cite{zhang2023openood}.
Also worth mentioning, there are several works that investigate adversarial vulnerabilities and robustness of OOD detectors \cite{fort2022adversarial,chen2022robust, ChenLWLJ21, yoon2022adversarial, BitterwolfM020, SehwagBSSCCM19, ibrahim2023outofdistribution, 9361176}
However, these works all utilize non-foundational underlying recognition models and consider very standard adversarial formulations involving classification labels. 
In short, these works do not consider the same adversarial threat model that we do.

\section{Attack Methodology} \label{sec:methods}


\begin{figure}[h]
    \centering
    \includegraphics[width=0.6\textwidth]{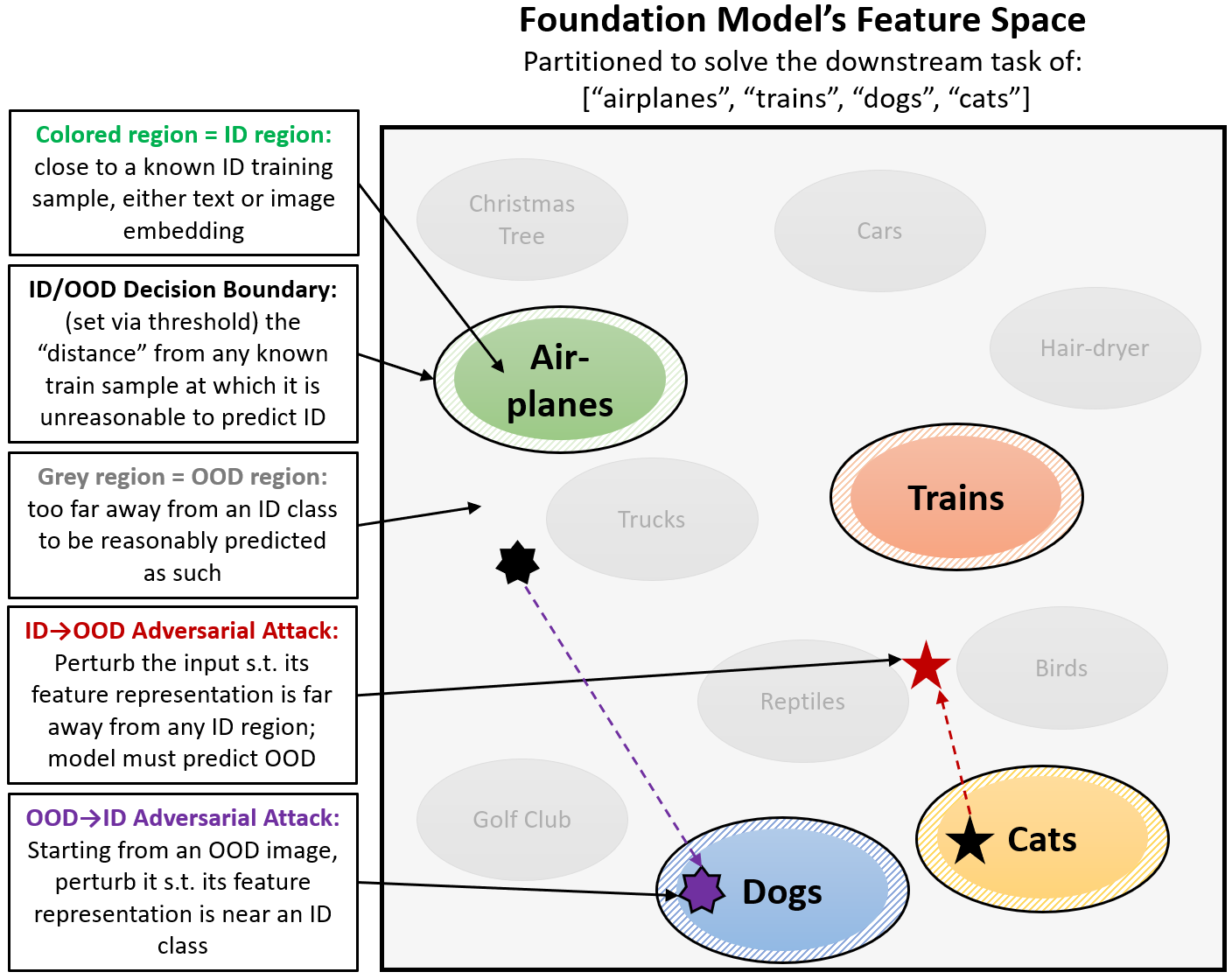}
    \caption{Visual representation of our attack concepts in a foundation model's feature space which has been partitioned to solve a specific downstream task.}
    \label{fig:attack_concept}
\end{figure}

Figure~\ref{fig:attack_concept} visualizes our attack methodology from a feature space perspective.
The main box represents a foundation model's feature space (e.g., CLIP or DINOv2) which has been partitioned to solve a downstream recognition task of [``Airplanes,'' ``Trains,'' ``Dogs,'' ``Cats''].
Because this feature extractor is so expressive off-the-shelf, it can solve this downstream task without having to update the extractor weights, i.e., it does not have to be finetuned.
Note, these classes define the ID set for the task.

Critical to the motivation of this work, we posit that in order for this recognition model to have adequate robustness in an open-world downstream environment, one would have to use an OOD detector.
This is because the feature extractor is aware of many concepts from the open-world, so we must allow for an ``other''/OOD prediction to be made in the event that an observed test input does not belong to one of the task's classes.
When the model is being used in zero-shot, the recent MCM \cite{ming2022delving} OOD detector may be employed, and when a parameterized classifier head is used (e.g., in linear probe setting) one of the many post-hoc OOD detection techniques may be considered \cite{hendrycks2017a, liang2018enhancing, HendrycksBMZKMS22, zhang2023openood}.
See Figure~\ref{fig:zeroshot_vs_linprobe_overview} in the Appendix for a refresher on zero-shot vs linear probing.
Finally, given this model setup, our attacks work by perturbing images in a way that manipulates their feature space representations s.t. any classifier head being used on top of the (fixed) feature extractor would mis-identify the image.

\textbf{Notation:}
We now define some notation which will be used throughout the remainder of this work.
Our full notation follows the construction of a CLIP model, and can be repurposed to describe many other foundational models.
Let, $f_v$ and $p_v$ represent the vision encoder and vision projector, respectively.
Then, $f_t$ and $p_t$ represent the text encoder and text projector, respectively.
Since CLIP models have a shared feature space between text and images, $p_v(f_v(x))$ and $p_t(f_t(y))$ effectively refers to the same feature space, where $x$ is an image and $y$ is its corresponding text description.
Following \cite{ming2022delving}, we use $y=$``this is a photo of a $<$\texttt{target class}$>$'' here.
Note, for models like DINOv2 which only ship with a vision encoder we can still represent their function with $f_v$.


\subsection{ID$\rightarrow$OOD Attack}

Our first attack perturbs an ID input such that it's predicted by the target model as OOD.
We call our method the \textbf{Away From Start} (AFS) attack, and define it as
\begin{equation*}
    \min_{\delta \in \mathcal{S}} \mathrm{cos\_sim}(f_v(x+\delta), f_v(x)).
\end{equation*}
Here, $\mathrm{cos\_sim}$ means cosine similarity and $\mathcal{S}$ represents the allowable set of perturbations which we define via $\ell_\infty$ constraint \cite{madry2018towards}.
The intuition for AFS is to perturb the image $x$ with $\delta$ s.t.~its feature space representation $f_v(x+\delta)$ is far away from the clean image's representation $f_v(x)$, i.e., where it started.
By doing so, we are likely moving towards a lower probability region of feature space w.r.t.~the classification task.

A key feature of our method is the lack of assumptions required by the adversary in order to execute the attack.
The adversary does not need to know any specific classes in the target model's label space and does not need to know the target model's training data.
This method is also not specific to any architecture and can be applied to ViTs and CNNs, SimCLRs, DINO(v2)s, CLIPs, etc.
Besides impacting just the OOD detector, this method can also be thought of as a traditional ``untargeted'' attack; we are manipulating the representation that is input into the classifier head, so the class prediction is also expected to change (verified in Section~\ref{sec:experiments}).


\subsection{OOD$\rightarrow$ID Attack}

Our second attack perturbs an OOD input s.t.~it's predicted by the target model as one of the downstream task's ID classes.
We call this method the \textbf{Towards Target + Away From Start} attack, and define it as
\begin{equation*}
    \max_{\delta \in \mathcal{S}} \mathrm{cos\_sim}(p_v(f_v(x+\delta)), p_t(f_t(y))) - \lambda \mathrm{cos\_sim}(p_v(f_v(x+\delta)), p_v(f_v(x))).
\end{equation*}
Recall from Section~\ref{sec:introduction}, in this work we are performing a distal-style attack where the starting OOD image is random noise, so in this equation $x\sim \mathcal{U}(0,1)^{C \times H \times W}$ where $\mathcal{U}$ is the Uniform distribution.
$\lambda$ is simply a weighting term that balances the influence of the terms and is set empirically.
The intuition for the first term is to encourage the distal image to lie very near an ID class prototype in a CLIP feature space, which is obtained by embedding a text description of the target class $y$. The second term carries the same AFS intuition as before, and although is not conceptually required for this attack it has experimentally shown to contribute a useful regularization signal which boosts blackbox transferability~\cite{inkawhich_neurips20}.
Figure~\ref{fig:distal_samples} shows some example distals which are the result of this optimization.
We believe that within the ``rules'' of current s.o.t.a.~OOD detection benchmarks, these images would surely be considered OOD w.r.t.~an ID task trained on natural images.

This attack also requires very minimal assumptions on behalf of the adversary. 
Unlike many contemporary attacks, our adversary does not need to know the entire target model's/task's label space and instead only needs to know the name of a single (target) class.
Our adversary also does not need any sample images from the target class or from any pretraining dataset.
We consider it an important direction of future work to investigate the impact of different starting images, including natural images. 
We anticipate the semantic relationship of the starting image w.r.t.~the target class may have an impact on success and this experiment must be carefully designed.


\textbf{Optimization Tricks:}
The last detail of our methodology pertains to how we optimize these attacks. 
In the previous two sub-sections we defined attacking ``signals,'' but as many prior works have shown there are a handful of optimization tricks that can further boost attack performance and transferability, particularly in blackbox settings. 
In this work we use momentum \cite{mim_attack}, diverse inputs \cite{di_attack}, ensembles \cite{ensemble_attack1, ensemble_attack2}, and translation invariance \cite{ti_attack} methods when performing the optimization.
\section{Experiments} \label{sec:experiments}

This section contains all of the main experimental results for our work and is organized as follows.
Section~\ref{sec:idood_top} describes details related to our ID$\rightarrow$OOD attacks.
This includes setup information, results of attacking zero-shot target models, the impact of perturbation strength, results of attacking non-zero-shot models across training algorithms, and lastly some analyses to shed light on how the attack impacts feature space.
Section~\ref{sec:oodid_top} then describes setup and results of our OOD$\rightarrow$ID attacks. 
Our experiments closely follow the setup of those in \cite{ming2022delving}.
We consider four primary downstream ID tasks: \texttt{OxfordPets}, \texttt{Food101}, \texttt{ImageNet-20}, \texttt{ImageNet-100}; four primary sources of OOD: \texttt{iNaturalist}, \texttt{Places}, \texttt{SUN}, \texttt{Textures}; with all ID and OOD settings/splits copied from \cite{ming2022delving}.
The results of our reproducibility experiments for \cite{ming2022delving} are shown in Appendix Figure~\ref{fig:reproduce_MCM}.


\subsection{ID$\rightarrow$OOD Attacks} \label{sec:idood_top}

\subsubsection{Experimental Setup} \label{sec:idood_metrics}
We primarily use a model pool containing the following eight ``Zero-shot Image Classification'' models from HuggingFace: [\textit{openai/clip-vit-b-32}; \textit{openai/clip-vit-b-16}; \textit{openai/clip-vit-l-14}; \textit{laion/clip-vit-b-32}; \textit{laion/clip-vit-l-14}; \textit{laion/clip-vit-h-14}; \textit{laion/clip-convnext-L}; \textit{laion/clip-convnext-XXL}]. 
From Figure~\ref{fig:hf_downloads}, these are among the most popular models currently available.
Importantly, this pool contains both ViT \cite{DosovitskiyB0WZ21} and ConvNeXT \cite{convnext} backbones so we can investigate the impact of transferring attacks across backbone architectures.
Unless otherwise specified, our ID$\rightarrow$OOD attacks are all run with the following settings.
Within each task, we attack all ID inputs s.t.~there is one adversarial example for every clean ID image in the task's test dataset. 
All images are then input into all models in the pool.
We compute whitebox attack success on the target model that was used to generate the perturbation.
We then compute an average blackbox attack success over all models in the pool that are NOT used as whiteboxes.
Finally, we use $\ell_\infty~\epsilon=\sfrac{16}{255}$ as the default attack strength, 20 perturbation iterations in a PGD-style optimization, a momentum strength of $\mu=1.0$, and a diverse inputs policy of  min\_size=170, max\_size=224, transform\_prob=0.5.

\textbf{Measuring attack success.}
We use the following three metrics to measure attack success:
\begin{itemize}  
    \itemsep0em
    \item Accuracy ($\downarrow$) - the classification accuracy on the adversarially perturbed ID data (advID). The lower the accuracy, the more powerful the attack.
    \item AUROC ($\uparrow$) - Area under the receiver operating characteristic curve between cleanID (label=1) and advID (label=0). Serves as a measure of separability between the cleanID and advID OOD detection score distributions. A higher AUROC means higher separability, which indicates a more powerful attack.
    \item FNR95 ($\uparrow$) - The percentage of advID samples that fall below the 95\% TPR threshold computed on clean ID (a.k.a, false negative rate). This is also a measure of separability between the cleanID and advID OOD scores, where higher separability means a more powerful attack.
\end{itemize}


\subsubsection{Attacking Zero-Shot CLIP Models} \label{sec:idood_zshot}

\begin{table}[]
\caption{ID$\rightarrow$OOD attack potency on zero-shot target models ($\epsilon=\sfrac{16}{255}$).}
\label{tab:afs_attack}
\resizebox{1.\textwidth}{!}{
\begin{tabular}{lcccccccccccc}
\toprule
                                                                   & \multicolumn{6}{c}{\cellcolor{blue!10}Task = \texttt{OxfordPets}}                                                            & \multicolumn{6}{c}{\cellcolor{yellow!10}Task = 
                                                                   \texttt{Food101}}                                                             \\ 
                                                                   & \multicolumn{3}{c}{\begin{tabular}[c]{@{}c@{}}Whitebox\\ Attack Success\end{tabular}} & \multicolumn{3}{c}{\begin{tabular}[c]{@{}c@{}}Avg. Blackbox\\ Attack Success\end{tabular}} & \multicolumn{3}{c}{\begin{tabular}[c]{@{}c@{}}Whitebox\\ Attack Success\end{tabular}} & \multicolumn{3}{c}{\begin{tabular}[c]{@{}c@{}}Avg. Blackbox\\ Attack Success\end{tabular}} \\ \cmidrule(lr){2-4} \cmidrule(lr){5-7} \cmidrule(lr){8-10} \cmidrule(lr){11-13}
Whitebox Model(s)                                                  & acc$^\downarrow$         & auroc$^\uparrow$         & fnr95$^\uparrow$         & acc$^\downarrow$             & auroc$^\uparrow$           & fnr95$^\uparrow$           & acc$^\downarrow$         & auroc$^\uparrow$         & fnr95$^\uparrow$         & acc$^\downarrow$            & auroc$^\uparrow$          & fnr95$^\uparrow$          \\ \midrule
openai/clip-vit-b-32                                               & 1.7         & 96.3          & 85.0          & 49.9            & 93.4            & 68.0            & 0.0         & 95.8          & 76.9          & 17.8           & 96.8           & 85.6           \\
openai/clip-vit-b-16                                               & 1.2         & 95.1          & 76.8          & 46.1            & 93.9            & 71.3            & 0.0         & 98.0          & 90.6          & 13.1           & 96.9           & 85.5           \\
openai/clip-vit-l-14                                               & 7.0         & 98.6          & 93.5          & 57.6            & 89.1            & 53.3            & 0.2         & 97.6          & 87.8          & 19.7           & 94.6           & 76.5           \\
laion/clip-vit-b-32                                                & 1.0         & 93.5          & 69.3          & 31.6            & 96.9            & 84.2            & 0.0         & 97.1          & 84.1          & 13.6           & 98.1           & 91.8           \\
laion/clip-vit-l-14                                                & 0.5         & 98.9          & 96.5          & 19.4            & 97.6            & 88.4            & 0.0         & 99.1          & 96.7          & 7.2            & 97.9           & 90.3           \\
laion/clip-vit-h-14                                                & 0.2         & 99.3          & 98.7          & 16.4            & 97.9            & 89.8            & 0.0         & 99.6          & 98.9          & 5.5            & 98.2           & 91.3           \\ [0.2mm] \cdashline{1-13} \\[-2.7mm]
laion/clip-vit-b-32 + laion/clip-vit-h-14                          & -           & -             & -             & 7.9             & 99.1            & 96.2            & -           & -             & -             & 2.1            & 99.3           & 97.2           \\
laion/clip-vit-b-32 + laion/clip-vit-h-14 + dinov2\_vitb14         & -           & -             & -             & \textbf{4.0}             & \textbf{99.2}            & \textbf{96.8}            & -           & -             & -             & \textbf{1.5}            & \textbf{99.4}           & \textbf{97.4}           \\ \midrule
                                                                   & \multicolumn{6}{c}{\cellcolor{green!10}Task = \texttt{ImageNet-20}}                                                            & \multicolumn{6}{c}{\cellcolor{orange!10}Task = 
                                                                   \texttt{ImageNet-100}}                                                        \\ 
                                                                   & \multicolumn{3}{c}{\begin{tabular}[c]{@{}c@{}}Whitebox\\ Attack Success\end{tabular}} & \multicolumn{3}{c}{\begin{tabular}[c]{@{}c@{}}Avg. Blackbox\\ Attack Success\end{tabular}} & \multicolumn{3}{c}{\begin{tabular}[c]{@{}c@{}}Whitebox\\ Attack Success\end{tabular}} & \multicolumn{3}{c}{\begin{tabular}[c]{@{}c@{}}Avg. Blackbox\\ Attack Success\end{tabular}} \\ \cmidrule(lr){2-4} \cmidrule(lr){5-7} \cmidrule(lr){8-10} \cmidrule(lr){11-13}
Whitebox Model(s)                                                  & acc$^\downarrow$         & auroc$^\uparrow$         & fnr95$^\uparrow$         & acc$^\downarrow$             & auroc$^\uparrow$           & fnr95$^\uparrow$           & acc$^\downarrow$         & auroc$^\uparrow$         & fnr95$^\uparrow$         & acc$^\downarrow$            & auroc$^\uparrow$          & fnr95$^\uparrow$          \\ \midrule
openai/clip-vit-b-32                                               & 1.6         & 98.3          & 94.0          & 63.3            & 86.7            & 54.2            & 0.5         & 91.8          & 51.0          & 51.3           & 82.5           & 37.6           \\
openai/clip-vit-b-16                                               & 1.0         & 98.9          & 95.8          & 60.5            & 89.0            & 59.7            & 0.4         & 92.2          & 50.4          & 46.1           & 84.1           & 39.0           \\
openai/clip-vit-l-14                                               & 8.2         & 99.3          & 97.2          & 69.6            & 85.1            & 47.8            & 3.3         & 94.3          & 62.7          & 56.0           & 79.6           & 30.3           \\
laion/clip-vit-b-32                                                & 0.5         & 98.7          & 96.8          & 53.4            & 90.6            & 65.1            & 0.2         & 96.1          & 74.4          & 40.6           & 87.3           & 49.7           \\
laion/clip-vit-l-14                                                & 3.6         & 99.5          & 97.9          & 45.2            & 93.0            & 73.8            & 1.4         & 96.8          & 80.6          & 30.2           & 89.1           & 52.1           \\
laion/clip-vit-h-14                                                & 2.7         & 99.5          & 99.3          & 39.8            & 93.6            & 76.8            & 1.3         & 97.4          & 85.3          & 26.6           & 90.0           & 54.5           \\ [0.2mm] \cdashline{1-13} \\[-2.7mm]
laion/clip-vit-b-32 + laion/clip-vit-h-14                          & -           & -             & -             & 25.8            & 97.0            & 87.7            & -           & -             & -             & 15.3           & 93.6           & 66.6           \\
laion/clip-vit-b-32 + laion/clip-vit-h-14 + dinov2\_vitb14         & -           & -             & -             & \textbf{20.5}            & \textbf{97.9}            & \textbf{91.5}            & -           & -             & -             & \textbf{10.4}           & \textbf{94.4}           & \textbf{68.1}    \\ \bottomrule      
\end{tabular}
}
\end{table}

Table~\ref{tab:afs_attack} shows the results of the ID$\rightarrow$OOD AFS attack in the four unique downstream task environments when the target recognition model is operating in zero-shot.
Each row of the table uses different whitebox model(s) to generate adversarial samples.
Within each task the three attack success metrics are reported in both whitebox and blackbox settings\footnote{To be absolutely clear, whitebox means the adversary guessed correctly and is able to compute gradients on the target model directly. 
Blackbox means the adversarial samples are being transferred to a model that the adversary did not use when creating the perturbation.}.

From this table we highlight a few major results.
Most importantly, whitebox \textit{and} blackbox attacks are both devastating, as an adversary can cause misclassifications and false negatives from nearly any cleanID input using the AFS attack.
Second, blackbox transferability is dependent on whitebox (obviously), but \textit{laion/clip-vit-h-14} is reliably the best individual model to attack from, removing any guess-work. 
We believe this may relate to model size.
A more subtle related point is that cross-architecture blackbox transfers are powerful, as \textit{clip-convnext} models are in the blackbox pool for these ViT-based whiteboxes (see Figure~\ref{fig:itemized_idood} in Appendix for individual model transfer rates).
Our third result is that ensemble attacks significantly boost blackbox transferability (perhaps not surprisingly). 
The interesting observation here is that adding a DINOv2 model to the whitebox ensemble further boosts performance when attacking other CLIP models. 
To our knowledge, this cross-algorithm transferability has not been shown before.
Finally, we want to reiterate that this attack can be interpreted as an untargeted attack - even if the target model is not using an OOD detector our attacks still have a significant impact on accuracy.



\subsubsection{Analyzing the Impact of Perturbation Strength} \label{sec:idood_eps}

\begin{figure}[h]
    \centering
    \includegraphics[width=0.85\textwidth,trim={0 0 0 2.5cm},clip]{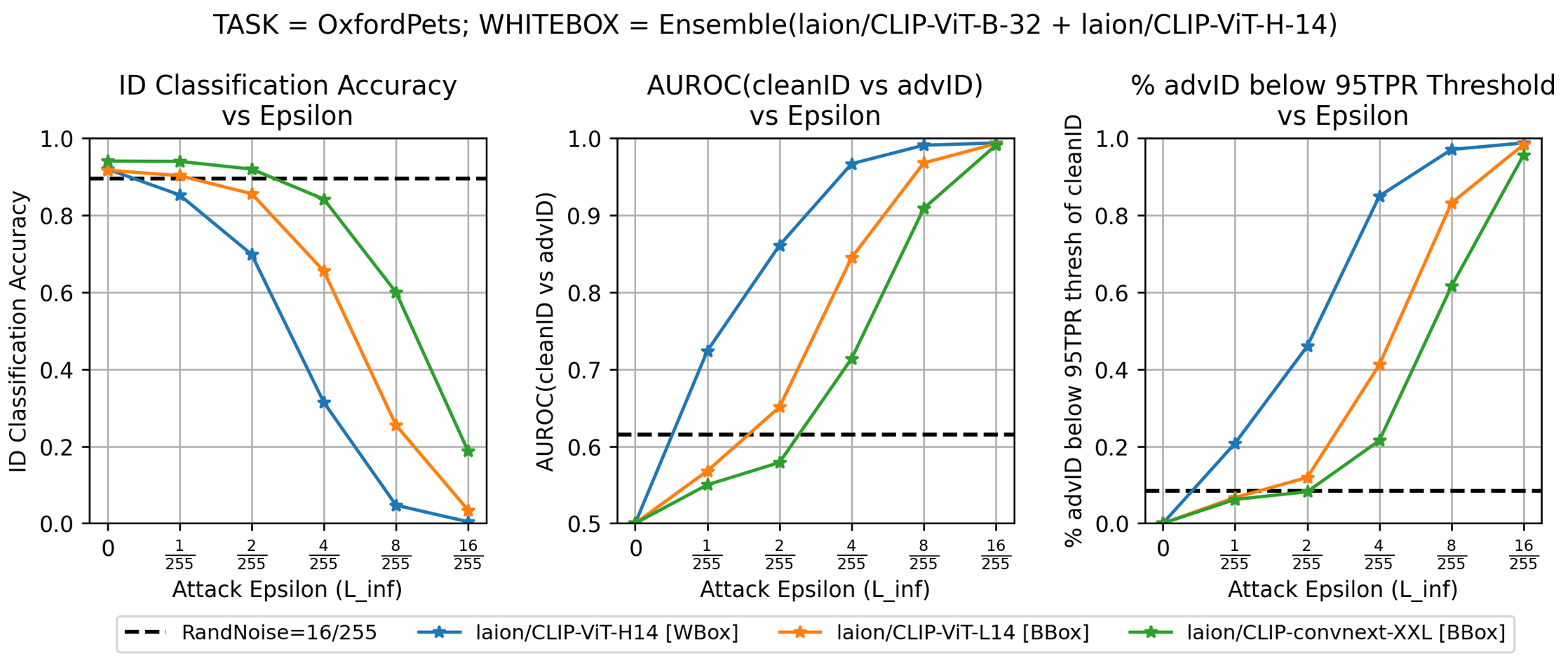}
    \caption{Impact of ID$\rightarrow$OOD attack at different perturbation strengths, i.e., $\ell_{\infty}~\epsilon$ values (task=\texttt{OxfordPets}; whitebox = \textit{laion/clip-vit-b-32} +  \textit{laion/clip-vit-h-14}).}
    \label{fig:attack_vs_eps}
\end{figure}


In Figure~\ref{fig:attack_vs_eps} we analyze the impact of using smaller perturbation strengths (i.e., $\epsilon$) on each of the attack success metrics.
Decreasing $\epsilon$ makes the perturbation even more imperceptible to a human observer (see Figure~\ref{fig:adv_examples} in Appendix) often at the cost of decreased attack potency.
For setup, we attack zero-shot target models solving the task of \texttt{OxfordPets} using a whitebox ensemble of: \textit{laion/clip-vit-b-32} + \textit{laion/clip-vit-h-14}.
We consider three different target models: {\color{NavyBlue}\textit{laion/clip-vit-h-14}},  {\color{BurntOrange}\textit{laion/clip-vit-l-14}}, and  {\color{ForestGreen}\textit{laion/clip-convnext-XXL}}, incurring whitebox, blackbox, and blackbox assumptions, respectively.
The x-axis of each plot is the $\ell_\infty$ attack $\epsilon$, where $\epsilon=0$ indicates no attack and serves as a point of reference.
As a sanity check, the black horizontal dashed lines show the average performance of the target models if we simply apply random Uniform noise to the test images at a strength of $\epsilon=\sfrac{16}{255}$.

The main trend worth noting here relates to the target models.
A far lower $\epsilon$ is required to attack the \textit{laion/clip-vit-h-14} model because it is in whitebox settings.
In fact, $\sfrac{\sim1}{2}$ the strength is often needed to achieve the same impact as the blackbox attacks at $\epsilon=\sfrac{16}{255}$.  
Between the two blackbox models, less $\epsilon$ is required to attack the \textit{laion/clip-vit-l-14} model which we believe is due to architectural similarity with the models in the ensemble (i.e., it's a ViT). 
The \textit{laion/clip-convnext-XXL} model uses a completely different backbone architecture and thus transferability lags.
To maximize adversarial success in the future, this would suggest having a diversity of model architectures in the attacking ensemble.
Finally, we observe the impact of random noise to be very small in comparison to the adversarial perturbations (verifying the potency of the adversary), and in all cases dropping to $\epsilon=\sfrac{8}{255}$ still has a devastating effect on any target model.


\subsubsection{Attacking non-CLIP and non-Zero-Shot Models}

In the previous two subsections all target models considered were CLIP models in zero-shot. 
In this section we attack models using different classification schemes (Linear Probing and Nearest Neighbors) as well as non-CLIP models (DINOv2 \cite{oquab2023dinov2} and SWAG \cite{singh2022revisiting}).
Note, the foundation models are still being used as a fixed feature extractors and the main difference is the instantiation of the classification heads because we now have empirical training images to support the task.
In this experiment we set the task to \texttt{OxfordPets} and by default the adversary uses a whitebox ensemble of: \textit{laion/clip-vit-b-32} + \textit{laion/clip-vit-h-14}.
Following \cite{radford21a}, we use scikit-learn for the Linear Probe \cite{sklearn_LR} and KNN \cite{sklearn_KNN} functions, which leverage the entire \texttt{OxfordPets} training dataset.

\begin{figure}[t]
    \centering
    \includegraphics[width=0.85\textwidth]{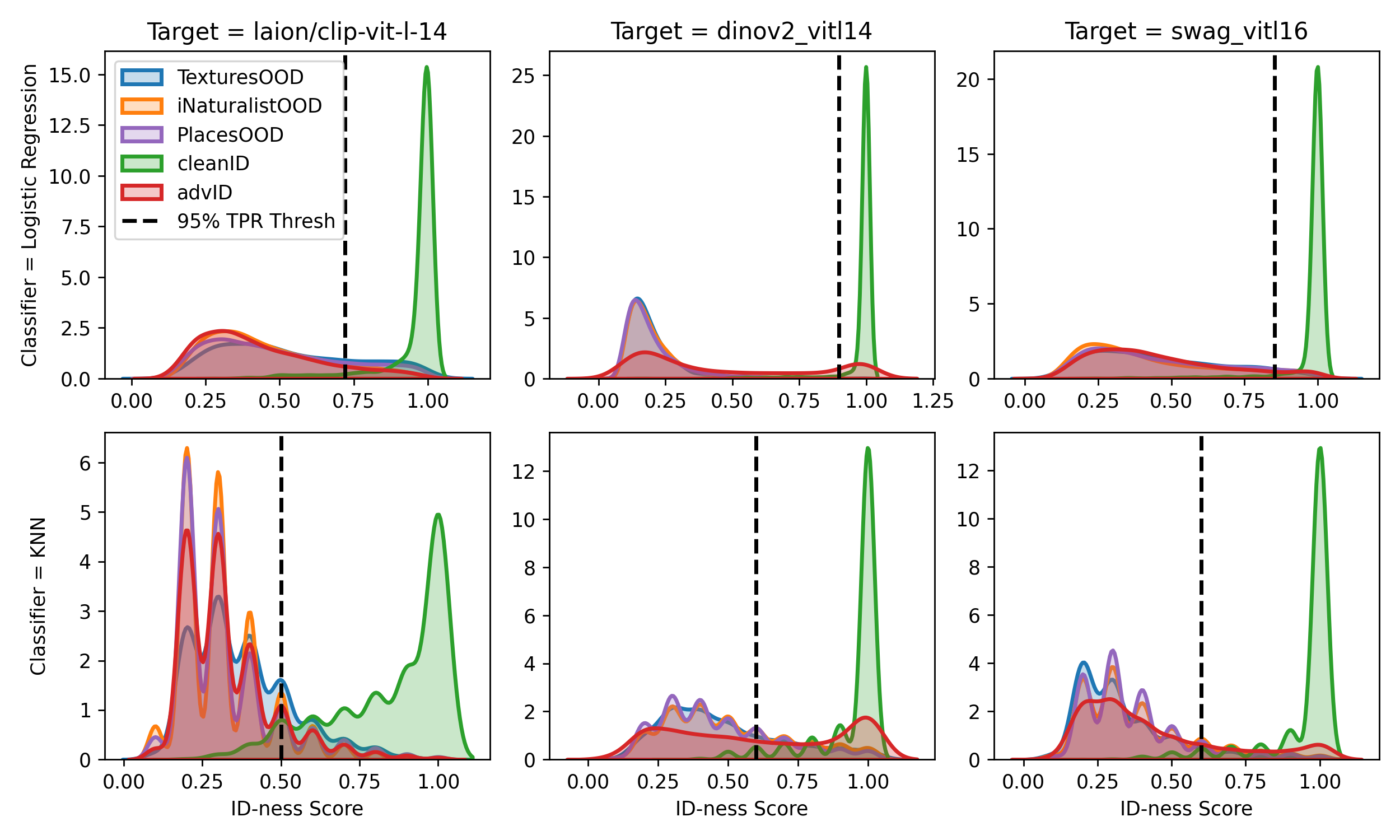}
    \caption{Impact of ID$\rightarrow$OOD attack when transferred to other target models (including DINOv2 and SWAG) using different types of classification heads (linear probing and KNN).}
    \label{fig:attack_knn_linprobe}
\end{figure}

Figure~\ref{fig:attack_knn_linprobe} displays the results of this experiment, where all of the target models are in blackbox settings. 
Each column corresponds to a different target model (from left to right: \textit{laion/clip-vit-l-14}, \textit{dinov2\_vitl14}, \textit{swag\_vitl16}) and each row is a different classification scheme (top=linear probe, bottom=KNN). 
The x-axis of each plot reflects the ``ID-ness'' score, where each is obtained using scikit-learn's \texttt{predict\_proba} function. 
We believe using this predicted probability score is equivalent to the very popular ``Max Softmax Probability'' (MSP) score from \cite{hendrycks2017a} and gives a reasonable estimate of OOD detection ability.
Finally, these plots can be interpreted identically to Figure~\ref{fig:attack_intuition}~(left).

The important takeaway from this figure is that our attack, as generated on an ensemble of two CLIP models, is able to significantly impact the OOD detection ability of all of these models, including DINOv2 and SWAG!
Regardless of backbone model or classifier head, our attack makes the adversarially perturbed ID data (advID) nearly indistinguishable from the naturally occurring OOD data (TexturesOOD, iNaturalistOOD, PlacesOOD).
Supplementing the findings from Section~\ref{sec:idood_zshot}, to our knowledge we are the first to observe such a transferability across foundational model types, which is especially interesting because the underlying learning mechanics of these algorithms are so different (e.g., CLIP learns with a contrastive signal between two modalities, DINOv2 learns via self-distillation from images only, and SWAG learns via weak-supervision).
We believe that this transferability suggests some amount of feature ``alignment'' between the different foundational models that is worth a future study.


\begin{table}[t]
\centering
\caption{ID$\rightarrow$OOD attack potency on non-CLIP target models (task=\texttt{OxfordPets}; $\epsilon=\sfrac{16}{255}$).}
\label{tab:cross_ssl}
\resizebox{1.\textwidth}{!}{
\begin{tabular}{lcccccccccccc}
\toprule
                                                           & \multicolumn{6}{c}{Target Model = dinov2\_vitl14}    & \multicolumn{6}{c}{Target Model = swag\_vitl16}      \\ \cmidrule(lr){2-7} \cmidrule(lr){8-13}
                                                           & \multicolumn{3}{c}{LogReg} & \multicolumn{3}{c}{kNN} & \multicolumn{3}{c}{LogReg} & \multicolumn{3}{c}{kNN} \\ \cmidrule(lr){2-4} \cmidrule(lr){5-7} \cmidrule(lr){8-10} \cmidrule(lr){11-13}
Whitebox Model(s)                                                  & acc$^\downarrow$         & auroc$^\uparrow$         & fnr95$^\uparrow$         & acc$^\downarrow$             & auroc$^\uparrow$           & fnr95$^\uparrow$           & acc$^\downarrow$         & auroc$^\uparrow$         & fnr95$^\uparrow$         & acc$^\downarrow$            & auroc$^\uparrow$          & fnr95$^\uparrow$   \\ \midrule
laion/clip-vit-h-14 + laion/clip-vit-b-32                  & 38.1    & 95.8     & 79.2  & 41.0     & 72.3   & 47.8  & 11.0      & 98.1    & 91.5   & 11.6   & 91.1   & 76.7  \\
dinov2\_vitb14                                             & 3.2     & 98.7     & 92.7  & 3.7    & 87.2   & 57.8  & 14.7    & 96.2    & 75.9   & 16.1   & 78.2   & 44.4  \\
swag\_vitb16                                               & 70.3    & 88.0       & 45.9  & 71.7   & 46.2   & 14.4  & 10.6    & 97.2    & 83.9   & 11.1   & 83.8   & 58.1  \\ [0.2mm] \cdashline{1-13} \\[-2.7mm]
laion/clip-vit-h-14 + laion/clip-vit-b-32 + dinov2\_vitb14 & \textbf{2.5}     & \textbf{99.2}     & \textbf{95.5}  & \textbf{2.5}    & \textbf{89.6}   & \textbf{65.5}  & 4.9     & \textbf{98.4}    & \textbf{91.6}   & \textbf{4.9}    & \textbf{92.3}   & 74.2  \\
laion/clip-vit-h-14 + laion/clip-vit-b-32 + swag\_vitb16   & 30.6    & 97.2     & 84.6  & 32.7   & 78.2   & 51.6  & \textbf{4.8}     & 98.2    & \textbf{91.6}   & 5.2    & \textbf{92.3}   & \textbf{76.9} \\ \bottomrule
\end{tabular}
}
\end{table}

A complimentary experiment investigates the impact of changing the whitebox model(s) to simulate the adversary making good and bad guesses as to what the target model is.
Results are shown in Table~\ref{tab:cross_ssl}.
For reference, the numbers across the top row in this table (whitebox models = \textit{laion/clip-vit-b-32} + \textit{laion/clip-vit-h-14}) match the results in columns 2 and 3 of Figure~\ref{fig:attack_knn_linprobe}.
The key takeaway here is that mixed-algorithm ensembles are (always) better for transfer attacking.
For example, when targeting a \textit{dinov2\_vitl14}, it is more productive to use an ensemble with CLIPs and \textit{dinov2\_b14} than it is to simply use the \textit{dinov2\_b14}. 
The same goes for the SWAG target model and the same was observed in Table~\ref{tab:afs_attack}.
Practically, this result should boost our concern regarding the feasibility of an adversarial attack when using s.o.t.a.~foundational models - the adversary may not have to ``try that hard'' to attack a variety of today's most powerful models, and instead simply construct an ensemble of diverse and popular ones they see on HuggingFace and \texttt{torch.hub}.


\subsubsection{Analyzing Attack Behaviors}

\begin{figure}[t]
    \centering
    \includegraphics[width=0.85\textwidth]{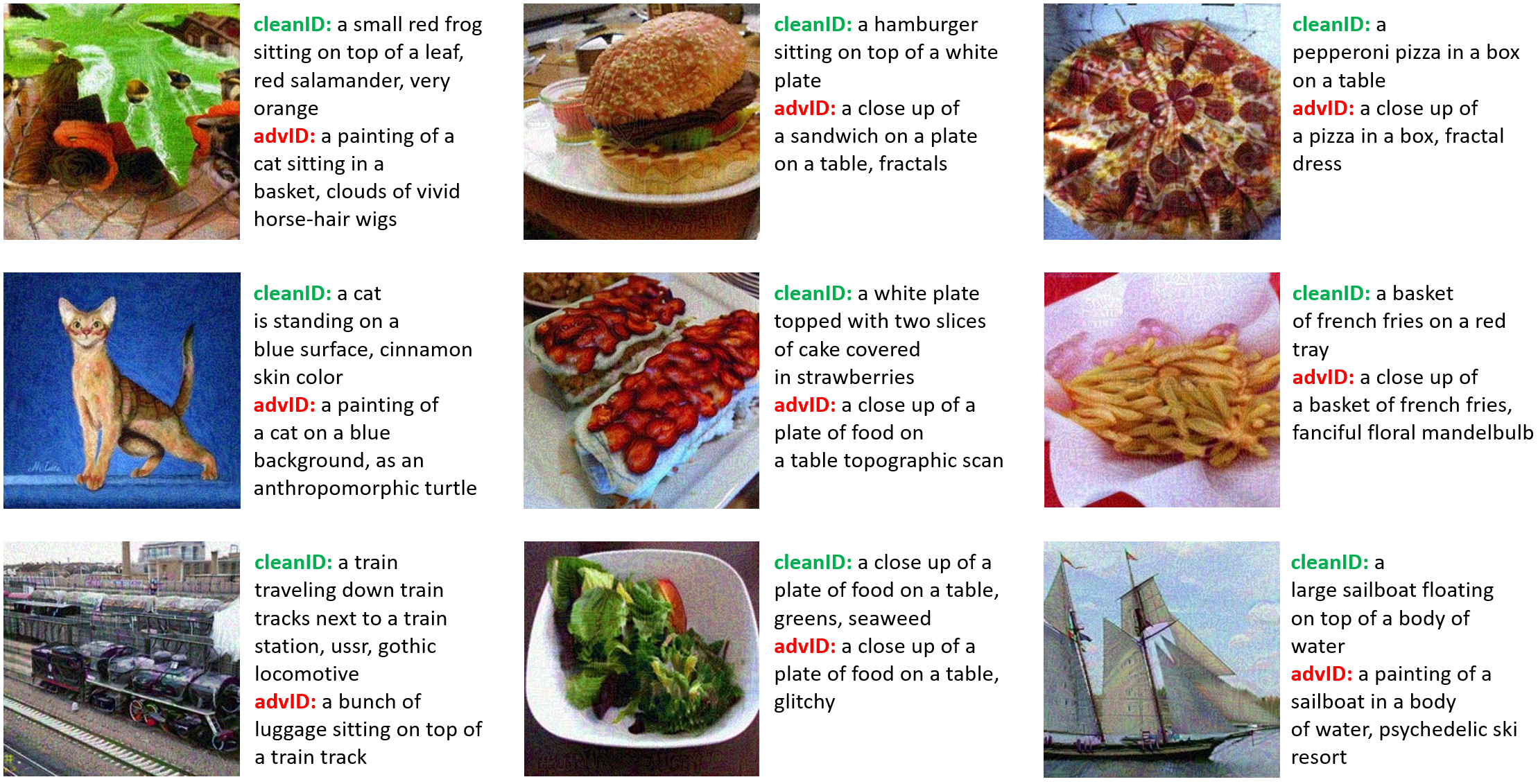}
    \caption{CLIP-Interrogator \cite{clip_interrogator} results on clean and adversarial samples.}
    \label{fig:interrogations}
\end{figure}

\textbf{Interrogations:}
In this section we perform two analyses to investigate how our attacks manipulate feature space from a different lens.
In the first study, shown in Figure \ref{fig:interrogations}, we use CLIP-Interrogator \cite{clip_interrogator} to inspect how the text prompts change between clean and adversarial images.
Functionally, we embed the images into the CLIP feature space then use CLIP-Interrogator to reverse engineer what the corresponding text prompt is.
From the figure we get some interesting results.
In several cases, like the red salamander and train images, the adversarial attack completely changes the semantics.
The adversarial salamander gets interrogated as a ``painting of a cat'' and the train as a ``bunch of luggage.''
However in most cases, the semantics mostly remain the same and the interrogator picks up on textural differences.
For example, the adversarial hamburger and pizzas are described as ``fractals,'' the sailboat and cat get tagged as ``paintings,'' and the cake and salad are associated with ``topographic scan'' and ``glitchy,'' respectively.
Also worth noting, in several cases the adversarial image's interrogation does not semantically change but gets more vague.
For example, hamburger becomes a sandwich and the strawberry cake becomes ``a plate of food on a table.''

This experiment highlights the challenge of task-specific OOD detection on top of open-world aware foundation models.
In this setting, developers of OOD detectors have to contend with a quasi-continuous feature space of these models as opposed to the binned and discrete feature space of supervised models.
Consider the case where an adversarially attacked hamburger becomes indistinguishable from a sandwich on a plate in feature space.
It is not technically incorrect to call a hamburger a sandwich, however, if the OOD detector threshold is willing to accommodate sandwich as ID w.r.t.~a hamburger support class, then it opens the model up to associating all other sandwiches (e.g., reubens) with hamburgers, which is not desirable in many cases.
A potential future direction to improve OOD detection here is to develop methods that embrace hierarchical classification which can respect a desired ontology \cite{linderman2022finegrain}.

\begin{figure}[t]
    \centering
    \includegraphics[width=0.80\textwidth]{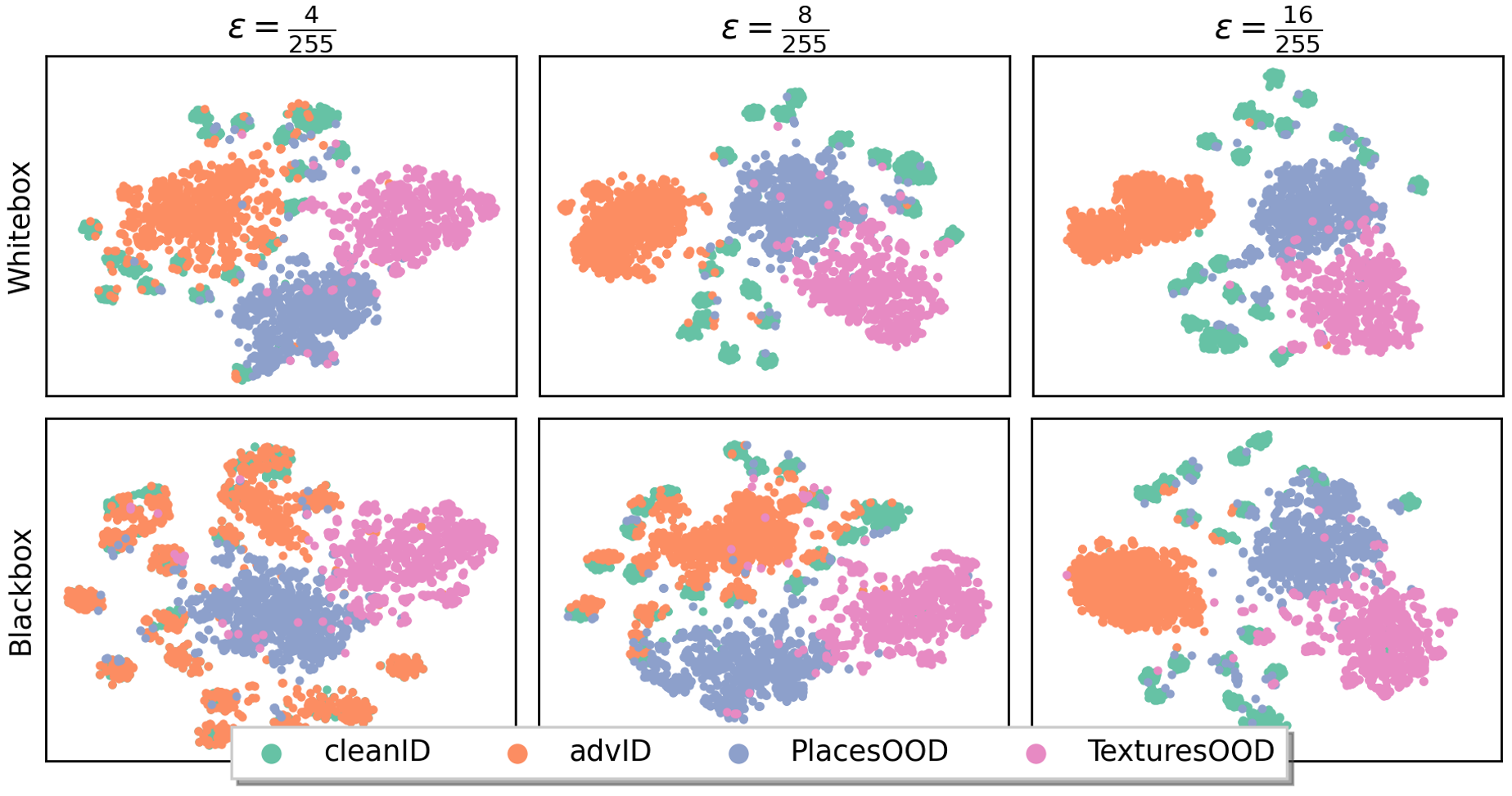}
    \caption{t-SNE plot showing how adversarial attacks manipulate the CLIP features space.}
    \label{fig:tsne}
\end{figure}

\textbf{Visualizing Feature Space:}
The second analysis visualizes the attacked CLIP feature space with t-SNE.
For setup, we use a \textit{laion/clip-vit-h-14} whitebox and \textit{openai/clip-vit-l-14} blackbox with data from the task of \texttt{ImageNet-20}.
The top row of Figure~\ref{fig:tsne} shows the whitebox model's feature space and the bottom shows the blackbox's.
Each column employs a different AFS attack $\epsilon$.
A clear takeaway is how the representations of the advID data change with $\epsilon$. 
At small $\epsilon=\sfrac{4}{255}$, the advID points are mixed among the cleanID data and are mostly distinct from the naturally occurring OOD.
However, as $\epsilon$ increases to $\sfrac{8}{255}$ in the whitebox and $\sfrac{16}{255}$ in the blackbox the advID points become completely separated from the cleanID data and live in their own subspace which appears to be organized as a single cluster.
At this point, they are virtually indistinguishable from the naturally occurring outliers and may not even be separable by class.
Overall, this plot echoes the intuition gained in previous experiments (particularly Section~\ref{sec:idood_eps}) and highlights how/why our attack is potent across classification schemes using a fixed feature extractor.

\subsection{OOD$\rightarrow$ID Attacks} \label{sec:oodid_top}

\subsubsection{Experimental Setup}

\begin{wrapfigure}{r}{.32\linewidth}
    \vspace{-15mm}
    \centering
    \includegraphics[width=0.99\linewidth]{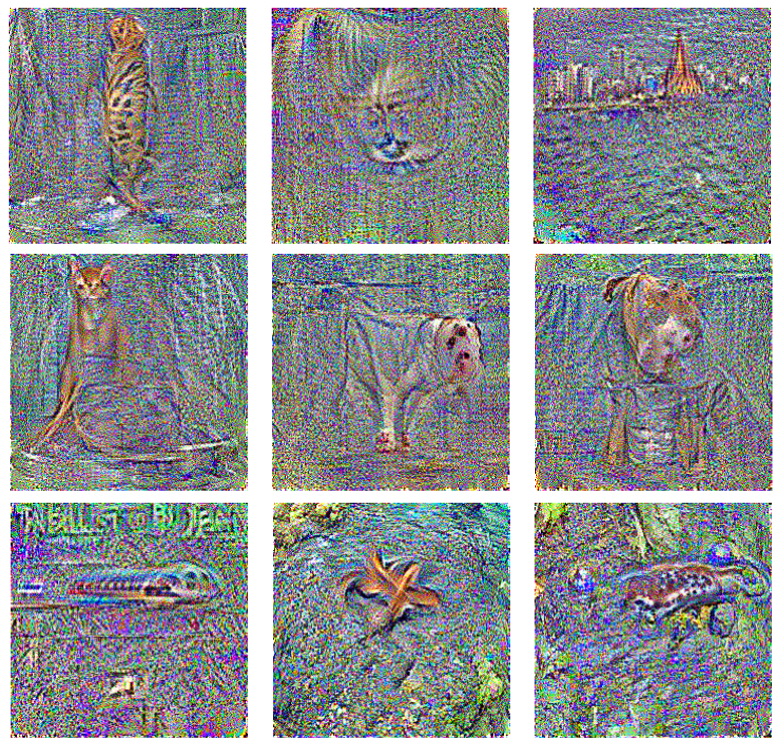}
    \caption{Distal samples generated with OOD$\rightarrow$ID attack.}
    \vspace{-5mm}
    \label{fig:distal_samples}
\end{wrapfigure}

To test the OOD$\rightarrow$ID attack we use some of the same models from the previous section.
Our adversary employs a whitebox ensemble of \textit{laion/clip-vit-b-32} + \textit{laion/clip-vit-h-14}.
The designated whitebox model for reporting attack success is \textit{laion/clip-vit-h-14}.
The designated blackbox model for reporting attack success is \textit{openai/clip-vit-l-14} (the most popular Zero-shot model on HuggingFace, ref. Figure~\ref{fig:hf_downloads}).
We set $\lambda$=0.25 in the Towards Target + Away From Start formulation.
Related to optimization, we use 500 PGD-style iterations, a momentum strength of $\mu=1.0$, a diverse inputs policy of  min\_size=170, max\_size=224, transform\_prob=0.5, and a translation invariant kernel\_size=5.
Finally, all results are averaged over 1000 distals balanced over possible target classes.
Figure~\ref{fig:distal_samples} shows some of the generated distal samples for classes in \texttt{OxfordPets} and \texttt{ImageNet-20}. 
To reiterate, although there are some structured features present in these images, we argue that within the rules of current OOD detection work these would undoubtedly be considered OOD w.r.t. the natural imagery domain.

\textbf{Measuring attack success:}
We use the following three metrics to measure attack success:
\begin{itemize}  
    \itemsep0em
    \item Targeted Success Rate (tSuc, $\uparrow$) - the percentage of distal images that are predicted by the target model as the chosen target class. Higher tSuc indicates more powerful attack.
    \item AUROC ($\downarrow$) - Area under the receiver operating characteristic curve between cleanID (label=1) and distals (label=0). In this case, a lower AUROC ($\leq0.5$) indicates a more powerful attack because it means that the distal OOD scores are either indistinguishable from, or higher than the cleanID scores.
    \item FPR95 ($\uparrow$) - The percentage of distal samples that fall above the 95\% TPR threshold of cleanID (a.k.a., false positive rate). This is also an indicator of how easy it is to distinguish cleanID from distals, where higher FPR95 signals a more powerful attack.
\end{itemize}

\begin{table}[t]
\centering
\caption{OOD$\rightarrow$ID distal attack potency.}
\label{tab:distal_attack}
\resizebox{0.6\textwidth}{!}{
\begin{tabular}{lcccccc}
\toprule
 & \multicolumn{3}{c}{Whitebox} & \multicolumn{3}{c}{Blackbox} \\ \cmidrule(lr){2-4} \cmidrule(lr){5-7}
Target Model's ID Task           & tSuc$^\uparrow$     & auroc$^\downarrow$   & fpr95$^\uparrow$   & tSuc$^\uparrow$    & auroc$^\downarrow$    & fpr95$^\uparrow$   \\ \midrule
\texttt{OxfordPets}                                                  & 100.0    & 0.2     & 100.0   & 77.8    & 79.6     & 69.4    \\
\texttt{ImageNet-20}                                                  & 100.0    & 0       & 100.0   & 94.0    & 71.6     & 82.6    \\ \bottomrule
\end{tabular}
}
\end{table}

\begin{figure}[t]
    \centering
    \includegraphics[width=1.0\textwidth]{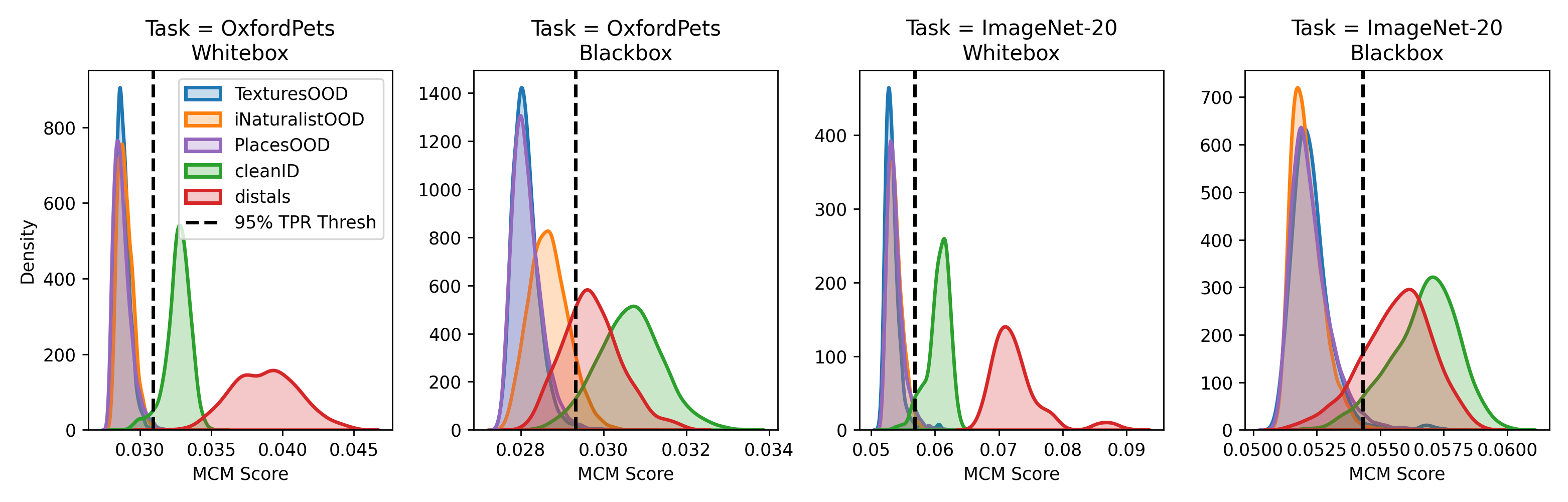}
    \caption{Main OOD$\rightarrow$ID attack results on whitebox and blackbox models.}
    \label{fig:distal_results}
\end{figure}


\subsubsection{Results}

Figure~\ref{fig:distal_results} displays the main results of our experiment (same interpretation as Figure~\ref{fig:attack_intuition}~(right)) and is supplemented by Table~\ref{tab:distal_attack}.
The left two subplots show attacks on target models performing zero-shot classification of \texttt{OxfordPets} and the right two subplots are target models performing zero-shot classification of \texttt{ImageNet-20}, all of which are using MCM OOD detectors.
The attacker has whitebox access to models in the $1^{st}$ and $3^{rd}$ columns and blackbox access to models in the $2^{nd}$ and $4^{th}$. 

It's immediately clear that distals are very powerful forms of attack.
In whitebox settings, distals can have even higher MCM scores (ID-ness) than the cleanID natural images, causing $\sim$0.0 AUROC, 100\% FPR95, and 100\% tSuc.
In blackbox transfer settings, distals remain potent and across both of the tasks a majority of distals would have been predicted by the blackbox model as ID \textit{and} as the attacker's chosen target class (tSuc~$\geq$~77.8\%).
We leave it to future work to explore the power of our Towards Target + Away From Start attack when the starting image is from the natural imagery domain.
We also encourage future work to tweak the threat model s.t.~the attacker is granted slightly more information about the downstream task to make the attack more powerful (e.g., some images or additional class names for enhanced context).

\section{Conclusion and Future Work}

Foundational vision models are rapidly improving and are becoming very useful even without finetuning on downstream tasks \cite{oquab2023dinov2}.
Current generation models like CLIP and DINOv2 have many improved properties over traditional supervised learning models like $\uparrow$ flexibility, faster task solving, and the ability to function with little-to-no labeled data.
The existence of good OOD detectors like MCM even makes them seem ``robust.''
It is our belief that these models will change the way the community thinks about training, using, and deploying deep learning models.
However, prohibitive training costs makes their distribution centralized and their usage somewhat predictable, creating patterns that a crafty adversary can exploit.

Our goal in this work is to expose some of these adversarial vulnerabilities in an effort to make future models more robust.
Specifically, we show that current foundational vision models are highly vulnerable to simple adversaries in both whitebox and blackbox settings.
We design attacks that require minimal knowledge assumptions about how the model is being used downstream.
Our methods can reliably make ID images ``look'' OOD to the model (i.e., create false negatives) as well as make garbage OOD inputs appear ID (i.e., inject false positives).
In a first of its kind experiment, we also show that adversarial perturbations crafted on CLIP models can transfer across algorithms to DINOv2 and SWAG-based models.
Finally, we show our attacks to be effective against several common classification schemes (including zero-shot, linear probing, and KNN) and provide analyses that illustrate how our adversaries manipulate the feature space.

As this work is likely only the beginning of a long journey towards robust foundational models, the following are some suggestions for future work inspired by what we observed here:
\begin{itemize}
    \itemsep0em
    \item Research to define policies for how to best/most-robustly use public foundational models under different amounts of compute resources. 
    For example, if compute is not a problem for an end user maybe some adversarial training-based finetuning makes sense. 
    However, one must be careful to control the trade-off between improved robustness and lower flexibility/accuracy. 
    On the other hand, if compute is a problem for the end user, what can be done?
    \item Investigation into adversarially robust representation learning at the foundational model scale which is \textbf{not} strictly limited to $\ell_p$-constrained adversaries.
    \item Develop improved attacking methods for current-generation foundational vision models, with an emphasis on low-assumption threat models. The more holes we find now, the more robust future models will be. Also along this line, it would be interesting to expose vulnerabilities of models like Segment-Anything and OWLv2 which contain localization components.
    \item In-depth study into how/why cross-algorithm transferability works so well. We were surprised that adversarial samples generated on CLIP models transferred to DINOv2 and SWAG models despite the learning algorithms being so different, and would be interested to understand this phenomenon. This could potentially expand the convergent learning theories from efforts focused on supervised learning \cite{LiYCLH15}.
\end{itemize}

\noindent
\textbf{Disclaimer:} The views expressed in this article are those of the authors and do not reflect official policy of the United States Air Force, Department of Defense or the U.S.~Government. Public release number: AFRL-2023-3382.

\printbibliography

@InProceedings{radford21a,
  title = 	 {Learning Transferable Visual Models From Natural Language Supervision},
  author =       {Radford, Alec and Kim, Jong Wook and Hallacy, Chris and Ramesh, Aditya and Goh, Gabriel and Agarwal, Sandhini and Sastry, Girish and Askell, Amanda and Mishkin, Pamela and Clark, Jack and Krueger, Gretchen and Sutskever, Ilya},
  booktitle = 	 {International Conference on Machine Learning ({ICML})},
  year = 	 {2021},
}

@misc{oquab2023dinov2,
      title={DINOv2: Learning Robust Visual Features without Supervision}, 
      author={Maxime Oquab and Timothée Darcet and Théo Moutakanni and Huy Vo and Marc Szafraniec and Vasil Khalidov and Pierre Fernandez and Daniel Haziza and Francisco Massa and Alaaeldin El-Nouby and Mahmoud Assran and Nicolas Ballas and Wojciech Galuba and Russell Howes and Po-Yao Huang and Shang-Wen Li and Ishan Misra and Michael Rabbat and Vasu Sharma and Gabriel Synnaeve and Hu Xu and Hervé Jegou and Julien Mairal and Patrick Labatut and Armand Joulin and Piotr Bojanowski},
      year={2023},
      eprint={2304.07193},
      archivePrefix={arXiv},
}

@inproceedings{singh2022revisiting,
      title={{Revisiting Weakly Supervised Pre-Training of Visual Perception Models}}, 
      author={Singh, Mannat and Gustafson, Laura and Adcock, Aaron and Reis, Vinicius de Freitas and Gedik, Bugra and Kosaraju, Raj Prateek and Mahajan, Dhruv and Girshick, Ross and Doll{\'a}r, Piotr and van der Maaten, Laurens},
      booktitle={IEEE Computer Vision and Pattern Recognition Conference ({CVPR})},
      year={2022}
}

@inproceedings{ming2022delving,
  title={Delving into Out-of-Distribution Detection with Vision-Language Representations},
  author={Ming, Yifei and Cai, Ziyang and Gu, Jiuxiang and Sun, Yiyou and Li, Wei and Li, Yixuan},
  booktitle={Advances in Neural Information Processing Systems ({NeurIPS})},
  year={2022}
}

@inproceedings{inkawhich_cvpr19,
  author    = {Nathan Inkawhich and
               Wei Wen and
               Hai Li and
               Yiran Chen},
  title     = {Feature Space Perturbations Yield More Transferable Adversarial Examples},
  booktitle = {IEEE Computer Vision and Pattern Recognition Conference ({CVPR})},
  year      = {2019}
}

@inproceedings{inkawhich_iclr20,
  author    = {Nathan Inkawhich and
               Kevin J Liang and
               Lawrence Carin and
               Yiran Chen},
  title     = {Transferable Perturbations of Deep Feature Distributions},
  booktitle = {International Conference on Learning Representations ({ICLR})},
  year      = {2020}
}

@inproceedings{inkawhich_neurips20,
  author    = {Nathan Inkawhich and
               Kevin J Liang and
               Binghui Wang and
               Matthew Inkawhich and
               Lawrence Carin and
               Yiran Chen},
  title     = {Perturbing Across the Feature Hierarchy to Improve Standard and Strict
               Blackbox Attack Transferability},
  booktitle = {Advances in Neural Information Processing Systems ({NeurIPS})},
  year      = {2020}
}

@inproceedings{DosovitskiyB0WZ21,
  author       = {Alexey Dosovitskiy and
                  Lucas Beyer and
                  Alexander Kolesnikov and
                  Dirk Weissenborn and
                  Xiaohua Zhai and
                  Thomas Unterthiner and
                  Mostafa Dehghani and
                  Matthias Minderer and
                  Georg Heigold and
                  Sylvain Gelly and
                  Jakob Uszkoreit and
                  Neil Houlsby},
  title        = {An Image is Worth 16x16 Words: Transformers for Image Recognition
                  at Scale},
  booktitle    = {International Conference on Learning Representations ({ICLR})},
  year         = {2021}
}

@inproceedings{tian2023designing,
  author  = {Keyu Tian and Yi Jiang and Qishuai Diao and Chen Lin and Liwei Wang and Zehuan Yuan},
  title   = {Designing BERT for Convolutional Networks: Sparse and Hierarchical Masked Modeling},
  booktitle  = {International Conference on Learning Representations ({ICLR})},
  year    = {2023},
}

@inproceedings{convnext,
  author       = {Zhuang Liu and
                  Hanzi Mao and
                  Chao{-}Yuan Wu and
                  Christoph Feichtenhofer and
                  Trevor Darrell and
                  Saining Xie},
  title        = {A ConvNet for the 2020s},
  booktitle    = {IEEE Computer Vision and Pattern Recognition Conference ({CVPR})},
  year         = {2022}
}

@inproceedings{ensemble_attack1,
  author    = {Yanpei Liu and
               Xinyun Chen and
               Chang Liu and
               Dawn Song},
  title     = {Delving into Transferable Adversarial Examples and Black-box Attacks},
  booktitle = {International Conference on Learning Representations ({ICLR})},
  year      = {2017}
}

@inproceedings{ensemble_attack2,
  author    = {Florian Tram{\`{e}}r and
               Alexey Kurakin and
               Nicolas Papernot and
               Ian J. Goodfellow and
               Dan Boneh and
               Patrick D. McDaniel},
  title     = {Ensemble Adversarial Training: Attacks and Defenses},
  booktitle = {International Conference on Learning Representations ({ICLR})},
  year      = {2018}
}

@inproceedings{mim_attack,
  author    = {Yinpeng Dong and
               Fangzhou Liao and
               Tianyu Pang and
               Hang Su and
               Jun Zhu and
               Xiaolin Hu and
               Jianguo Li},
  title     = {Boosting Adversarial Attacks With Momentum},
  booktitle = {IEEE Computer Vision and Pattern Recognition Conference ({CVPR})},
  year      = {2018}
}

@inproceedings{di_attack,
  author    = {Cihang Xie and
               Zhishuai Zhang and
               Yuyin Zhou and
               Song Bai and
               Jianyu Wang and
               Zhou Ren and
               Alan L. Yuille},
  title     = {Improving Transferability of Adversarial Examples With Input Diversity},
  booktitle = {IEEE Computer Vision and Pattern Recognition Conference ({CVPR})},
  year      = {2019}
}

@inproceedings{ti_attack,
  author    = {Yinpeng Dong and
               Tianyu Pang and
               Hang Su and
               Jun Zhu},
  title     = {Evading Defenses to Transferable Adversarial Examples by Translation-Invariant
               Attacks},
  booktitle = {IEEE Computer Vision and Pattern Recognition Conference ({CVPR})},
  year      = {2019}
}

@misc{sklearn_LR,
  title = {Scikit-learn Logistic Regression Model},
  howpublished = {\url{https://scikit-learn.org/stable/modules/generated/sklearn.linear_model.LogisticRegression.html}},
  note = {Accessed: April 2023.}
}

@misc{sklearn_KNN,
  title = {Scikit-learn k-Nearest Neighbors},
  howpublished = {\url{https://scikit-learn.org/stable/modules/generated/sklearn.neighbors.KNeighborsClassifier.html}},
  note = {Accessed: April 2023.}
}

@misc{imagebind,
      title={ImageBind: One Embedding Space To Bind Them All}, 
      author={Rohit Girdhar and Alaaeldin El-Nouby and Zhuang Liu and Mannat Singh and Kalyan Vasudev Alwala and Armand Joulin and Ishan Misra},
      year={2023},
      eprint={2305.05665},
      archivePrefix={arXiv},
}

@inproceedings{BanD22,
  author       = {Yuanhao Ban and
                  Yinpeng Dong},
  title        = {Pre-trained Adversarial Perturbations},
  booktitle    = {Advances in Neural Information Processing Systems ({NeurIPS})},
  year         = {2022}
}

@misc{
sfort_clipattack2,
title={Pixels still beat text: Attacking the OpenAI CLIP model with text patches and adversarial pixel perturbations},
url={https://stanislavfort.github.io/2021/03/05/OpenAI_CLIP_stickers_and_adversarial_examples.html},
author={Stanislav Fort},
year={2021},
}

@misc{
sfort_clipattack1,
title={Adversarial examples for the OpenAI CLIP in its zero-shot classification regime and their semantic generalization},
url={https://stanislavfort.github.io/2021/01/12/OpenAI_CLIP_adversarial_examples.html},
author={Stanislav Fort},
year={2021},
}

@misc{noever_clipattack,
      title={Reading Isn't Believing: Adversarial Attacks On Multi-Modal Neurons}, 
      author={David A. Noever and Samantha E. Miller Noever},
      year={2021},
      eprint={2103.10480},
      archivePrefix={arXiv},
}

@inproceedings{mao_iclr23,
  author       = {Chengzhi Mao and
                  Scott Geng and
                  Junfeng Yang and
                  Xin Wang and
                  Carl Vondrick},
  title        = {Understanding Zero-Shot Adversarial Robustness for Large-Scale Models},
  booktitle = {International Conference on Learning Representations ({ICLR})},
  year         = {2023}
}

@misc{li2023languagedriven,
      title={Language-Driven Anchors for Zero-Shot Adversarial Robustness}, 
      author={Xiao Li and Wei Zhang and Yining Liu and Zhanhao Hu and Bo Zhang and Xiaolin Hu},
      year={2023},
      eprint={2301.13096},
      archivePrefix={arXiv},
}

@misc{
yoon2022adversarial,
title={Adversarial Distributions Against Out-of-Distribution Detectors},
author={Sangwoong Yoon and Jinwon Choi and Yonghyeon LEE and Yung-Kyun Noh and Frank C. Park},
year={2022},
}

@inproceedings{SehwagBSSCCM19,
  author       = {Vikash Sehwag and
                  Arjun Nitin Bhagoji and
                  Liwei Song and
                  Chawin Sitawarin and
                  Daniel Cullina and
                  Mung Chiang and
                  Prateek Mittal},
  title        = {Analyzing the Robustness of Open-World Machine Learning},
  booktitle    = {Artificial Intelligence and Security Conference ({AISec@CCS})},
  year         = {2019}
}

@misc{ibrahim2023outofdistribution,
      title={Towards Out-of-Distribution Adversarial Robustness}, 
      author={Adam Ibrahim and Charles Guille-Escuret and Ioannis Mitliagkas and Irina Rish and David Krueger and Pouya Bashivan},
      year={2023},
      eprint={2210.03150},
      archivePrefix={arXiv},
}

@inproceedings{BitterwolfM020,
  author       = {Julian Bitterwolf and
                  Alexander Meinke and
                  Matthias Hein},
  title        = {Certifiably Adversarially Robust Detection of Out-of-Distribution
                  Data},
  booktitle    = {Advances in Neural Information Processing Systems ({NeurIPS})},
  year         = {2020}
}

@misc{fort2022adversarial,
      title={Adversarial vulnerability of powerful near out-of-distribution detection}, 
      author={Stanislav Fort},
      year={2022},
      eprint={2201.07012},
      archivePrefix={arXiv},
}

@ARTICLE{9361176,
  author={Upadhyay, Ujjwal and Mukherjee, Prerana},
  journal={IEEE Signal Processing Letters}, 
  title={Generating Out of Distribution Adversarial Attack Using Latent Space Poisoning}, 
  year={2021},
  volume={28},
  number={},
  pages={523-527},
  doi={10.1109/LSP.2021.3061327}}

@inproceedings{
chen2022robust,
title={Robust Out-of-distribution Detection for Neural Networks},
author={Jiefeng Chen and Yixuan Li and Xi Wu and Yingyu Liang and Somesh Jha},
booktitle={The AAAI-22 Workshop on Adversarial Machine Learning and Beyond},
year={2022},
}

@inproceedings{ChenLWLJ21,
  author       = {Jiefeng Chen and
                  Yixuan Li and
                  Xi Wu and
                  Yingyu Liang and
                  Somesh Jha},
  title        = {{ATOM:} Robustifying Out-of-Distribution Detection Using Outlier Mining},
  booktitle    = {European Conference on Machine Learning and Principles and Practice of Knowledge Discovery in Databases ({ECML/PKDD})},
  year         = {2021}
}

@misc{carlini2023poisoning,
      title={Poisoning Web-Scale Training Datasets is Practical}, 
      author={Nicholas Carlini and Matthew Jagielski and Christopher A. Choquette-Choo and Daniel Paleka and Will Pearce and Hyrum Anderson and Andreas Terzis and Kurt Thomas and Florian Tramèr},
      year={2023},
      eprint={2302.10149},
      archivePrefix={arXiv},
}

@misc{kirillov2023segany,
      title={Segment Anything}, 
      author={Alexander Kirillov and Eric Mintun and Nikhila Ravi and Hanzi Mao and Chloe Rolland and Laura Gustafson and Tete Xiao and Spencer Whitehead and Alexander C. Berg and Wan-Yen Lo and Piotr Dollár and Ross Girshick},
      year={2023},
      eprint={2304.02643},
      archivePrefix={arXiv},
}

@inproceedings{ChenK0H20,
  author       = {Ting Chen and
                  Simon Kornblith and
                  Mohammad Norouzi and
                  Geoffrey E. Hinton},
  title        = {A Simple Framework for Contrastive Learning of Visual Representations},
  booktitle    = {International Conference on Machine Learning ({ICML})},
  year         = {2020}
}

@inproceedings{ZhangYS22,
  author       = {Jiaming Zhang and
                  Qi Yi and
                  Jitao Sang},
  title        = {Towards Adversarial Attack on Vision-Language Pre-training Models},
  booktitle    = {{ACM} Multimedia},
  year         = {2022}
}

@inproceedings{GrillSATRBDPGAP20,
  author       = {Jean{-}Bastien Grill and
                  Florian Strub and
                  Florent Altch{\'{e}} and
                  Corentin Tallec and
                  Pierre H. Richemond and
                  Elena Buchatskaya and
                  Carl Doersch and
                  Bernardo {\'{A}}vila Pires and
                  Zhaohan Guo and
                  Mohammad Gheshlaghi Azar and
                  Bilal Piot and
                  Koray Kavukcuoglu and
                  R{\'{e}}mi Munos and
                  Michal Valko},
  title        = {Bootstrap Your Own Latent - {A} New Approach to Self-Supervised Learning},
  booktitle    = {Advances in Neural Information Processing Systems ({NeurIPS})},
  year         = {2020}
}

@inproceedings{CaronTMJMBJ21,
  author       = {Mathilde Caron and
                  Hugo Touvron and
                  Ishan Misra and
                  Herv{\'{e}} J{\'{e}}gou and
                  Julien Mairal and
                  Piotr Bojanowski and
                  Armand Joulin},
  title        = {Emerging Properties in Self-Supervised Vision Transformers},
  booktitle    = {IEEE International Conference on Computer Vision ({ICCV})},
  year         = {2021}
}

@inproceedings{BardesPL22,
  author       = {Adrien Bardes and
                  Jean Ponce and
                  Yann LeCun},
  title        = {VICReg: Variance-Invariance-Covariance Regularization for Self-Supervised
                  Learning},
  booktitle    = {International Conference on Learning Representations ({ICLR})},
  year         = {2022}
}

@inproceedings{JiaYXCPPLSLD21,
  author       = {Chao Jia and
                  Yinfei Yang and
                  Ye Xia and
                  Yi{-}Ting Chen and
                  Zarana Parekh and
                  Hieu Pham and
                  Quoc V. Le and
                  Yun{-}Hsuan Sung and
                  Zhen Li and
                  Tom Duerig},
  title        = {Scaling Up Visual and Vision-Language Representation Learning With
                  Noisy Text Supervision},
  booktitle    = {International Conference on Machine Learning ({ICML})},
  year         = {2021}
}

@misc{ssl_cookbook,
      title={A Cookbook of Self-Supervised Learning}, 
      author={Randall Balestriero and Mark Ibrahim and Vlad Sobal and Ari Morcos and Shashank Shekhar and Tom Goldstein and Florian Bordes and Adrien Bardes and Gregoire Mialon and Yuandong Tian and Avi Schwarzschild and Andrew Gordon Wilson and Jonas Geiping and Quentin Garrido and Pierre Fernandez and Amir Bar and Hamed Pirsiavash and Yann LeCun and Micah Goldblum},
      year={2023},
      eprint={2304.12210},
      archivePrefix={arXiv},
}

@misc{abs-2306-09683,
      title={Scaling Open-Vocabulary Object Detection}, 
      author={Matthias Minderer and Alexey Gritsenko and Neil Houlsby},
      year={2023},
      eprint={2306.09683},
      archivePrefix={arXiv},
}

@inproceedings{
hendrycks2017a,
title={A Baseline for Detecting Misclassified and Out-of-Distribution Examples in Neural Networks},
author={Dan Hendrycks and Kevin Gimpel},
booktitle={International Conference on Learning Representations ({ICLR})},
year={2017},
}

@inproceedings{
liang2018enhancing,
title={Enhancing The Reliability of Out-of-distribution Image Detection in Neural Networks},
author={Shiyu Liang and Yixuan Li and R. Srikant},
booktitle={International Conference on Learning Representations ({ICLR})},
year={2018},
}

@misc{zhang2023openood,
      title={OpenOOD v1.5: Enhanced Benchmark for Out-of-Distribution Detection}, 
      author={Jingyang Zhang and Jingkang Yang and Pengyun Wang and Haoqi Wang and Yueqian Lin and Haoran Zhang and Yiyou Sun and Xuefeng Du and Kaiyang Zhou and Wayne Zhang and Yixuan Li and Ziwei Liu and Yiran Chen and Hai Li},
      year={2023},
      eprint={2306.09301},
      archivePrefix={arXiv},
}

@inproceedings{HendrycksBMZKMS22,
  author       = {Dan Hendrycks and
                  Steven Basart and
                  Mantas Mazeika and
                  Andy Zou and
                  Joseph Kwon and
                  Mohammadreza Mostajabi and
                  Jacob Steinhardt and
                  Dawn Song},
  title        = {Scaling Out-of-Distribution Detection for Real-World Settings},
  booktitle    = {International Conference on Machine Learning ({ICML})},
  year         = {2022}
}

@ARTICLE{9695222,
  author={Inkawhich, Nathan and Zhang, Jingyang and Davis, Eric K. and Luley, Ryan and Chen, Yiran},
  journal={IEEE Journal of Selected Topics in Applied Earth Observations and Remote Sensing}, 
  title={Improving Out-of-Distribution Detection by Learning From the Deployment Environment}, 
  year={2022},
  volume={15},
  number={},
  pages={2070-2086},
  doi={10.1109/JSTARS.2022.3146362}}

@inproceedings{Stutz0S20,
  author       = {David Stutz and
                  Matthias Hein and
                  Bernt Schiele},
  title        = {Confidence-Calibrated Adversarial Training: Generalizing to Unseen
                  Attacks},
  booktitle    = {International Conference on Machine Learning ({ICML})},
  year         = {2020}
}

@inproceedings{SzegedyZSBEGF13,
  author       = {Christian Szegedy and
                  Wojciech Zaremba and
                  Ilya Sutskever and
                  Joan Bruna and
                  Dumitru Erhan and
                  Ian J. Goodfellow and
                  Rob Fergus},
  title        = {Intriguing properties of neural networks},
  booktitle    = {International Conference on Learning Representations ({ICLR})},
  year         = {2014}
}

@inproceedings{GoodfellowSS14,
  author       = {Ian J. Goodfellow and
                  Jonathon Shlens and
                  Christian Szegedy},
  title        = {Explaining and Harnessing Adversarial Examples},
  booktitle    = {International Conference on Learning Representations ({ICLR})},
  year         = {2015}
}

@inproceedings{Carlini017,
  author       = {Nicholas Carlini and
                  David A. Wagner},
  title        = {Towards Evaluating the Robustness of Neural Networks},
  booktitle    = {{IEEE} Symposium on Security and Privacy},
  year         = {2017}
}

@inproceedings{InkawhichLZYLC21,
  author       = {Nathan Inkawhich and
                  Kevin J. Liang and
                  Jingyang Zhang and
                  Huanrui Yang and
                  Hai Li and
                  Yiran Chen},
  title        = {Can Targeted Adversarial Examples Transfer When the Source and Target
                  Models Have No Label Space Overlap?},
  booktitle    = {IEEE International Conference on Computer Vision Workshops ({ICCVW})},
  year         = {2021}
}

@inproceedings{
madry2018towards,
title={Towards Deep Learning Models Resistant to Adversarial Attacks},
author={Aleksander Madry and Aleksandar Makelov and Ludwig Schmidt and Dimitris Tsipras and Adrian Vladu},
booktitle={International Conference on Learning Representations ({ICLR})},
year={2018},
}

@misc{
clip_interrogator,
title={CLIP Interrogator},
url={https://github.com/pharmapsychotic/clip-interrogator},
author={pharmapsychotic},
year={2023},
}

@inproceedings{linderman2022finegrain,
      title={Fine-grain Inference on Out-of-Distribution Data with Hierarchical Classification}, 
      author={Randolph Linderman and Jingyang Zhang and Nathan Inkawhich and Hai Li and Yiran Chen},
      booktitle={Conference on Lifelong Learning Agents ({CoLLAs})},
      year={2023},
}

@inproceedings{LiYCLH15,
  author       = {Yixuan Li and
                  Jason Yosinski and
                  Jeff Clune and
                  Hod Lipson and
                  John E. Hopcroft},
  title        = {Convergent Learning: Do different neural networks learn the same representations?},
  booktitle    = {International Conference on Learning Representations ({ICLR})},
  year         = {2016}
}

\section*{Appendix}

\begin{figure}[h]
    \centering
    \includegraphics[width=1.\textwidth]{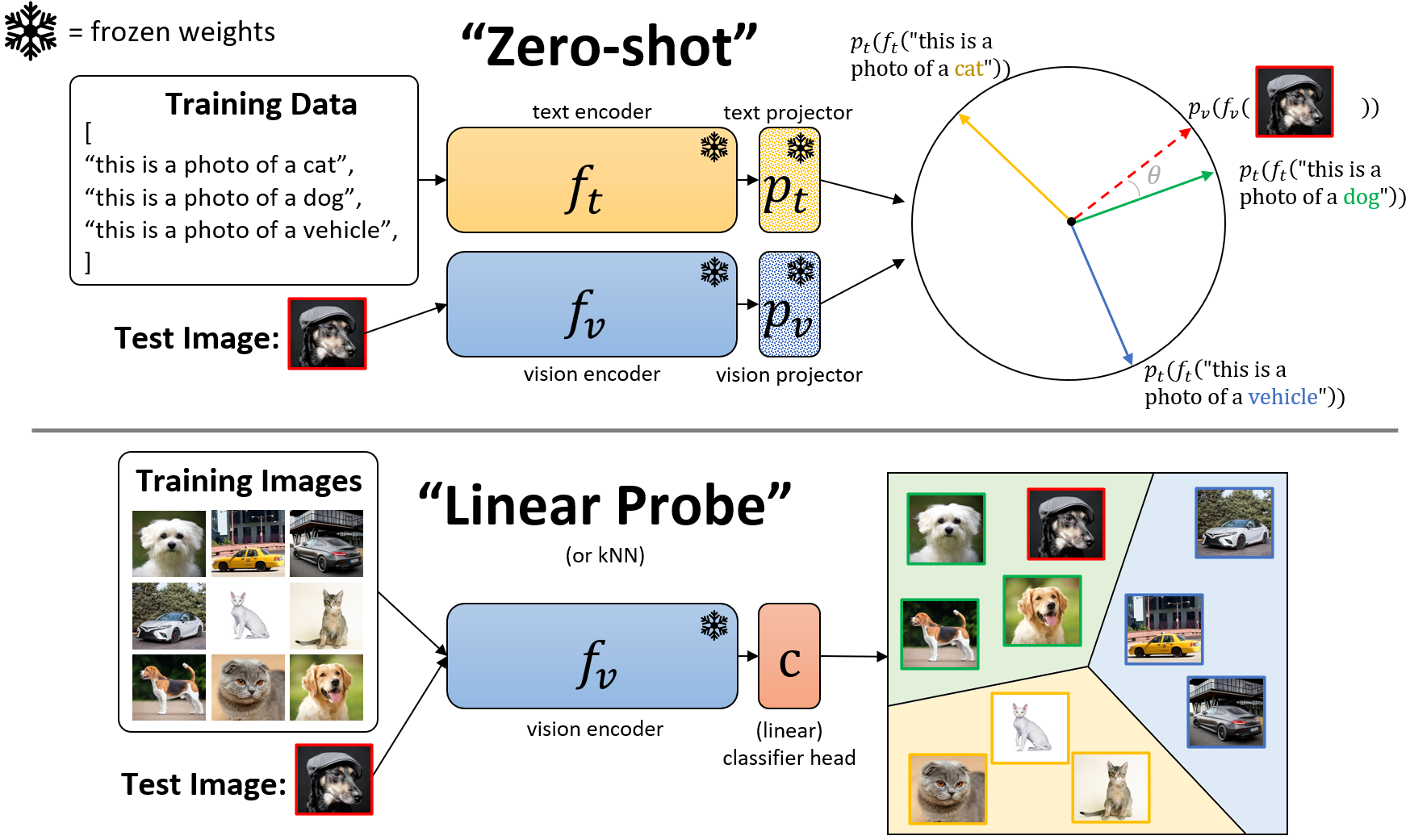}
    \caption{Overview of how foundational models are used in Zero-Shot and Linear Probe settings.}
    \label{fig:zeroshot_vs_linprobe_overview}
\end{figure}

\begin{figure}[h]
    \centering
    \includegraphics[width=1.\textwidth]{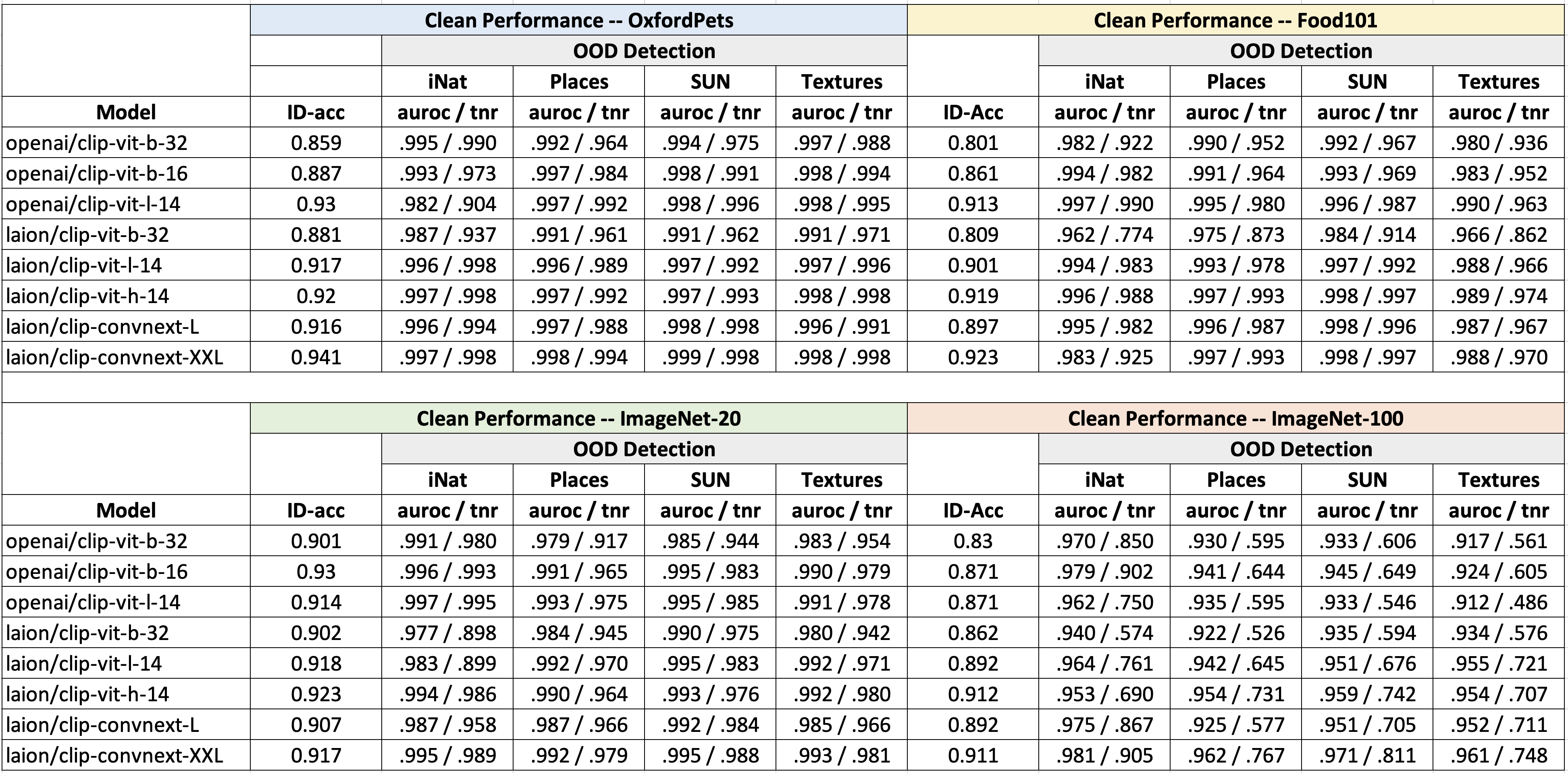}
    \caption{Reproducing MCM \cite{ming2022delving} OOD detection results with the datasets and models used in this work.}
    \label{fig:reproduce_MCM}
\end{figure}

\begin{figure}[h]
    \centering
    \includegraphics[width=1.\textwidth]{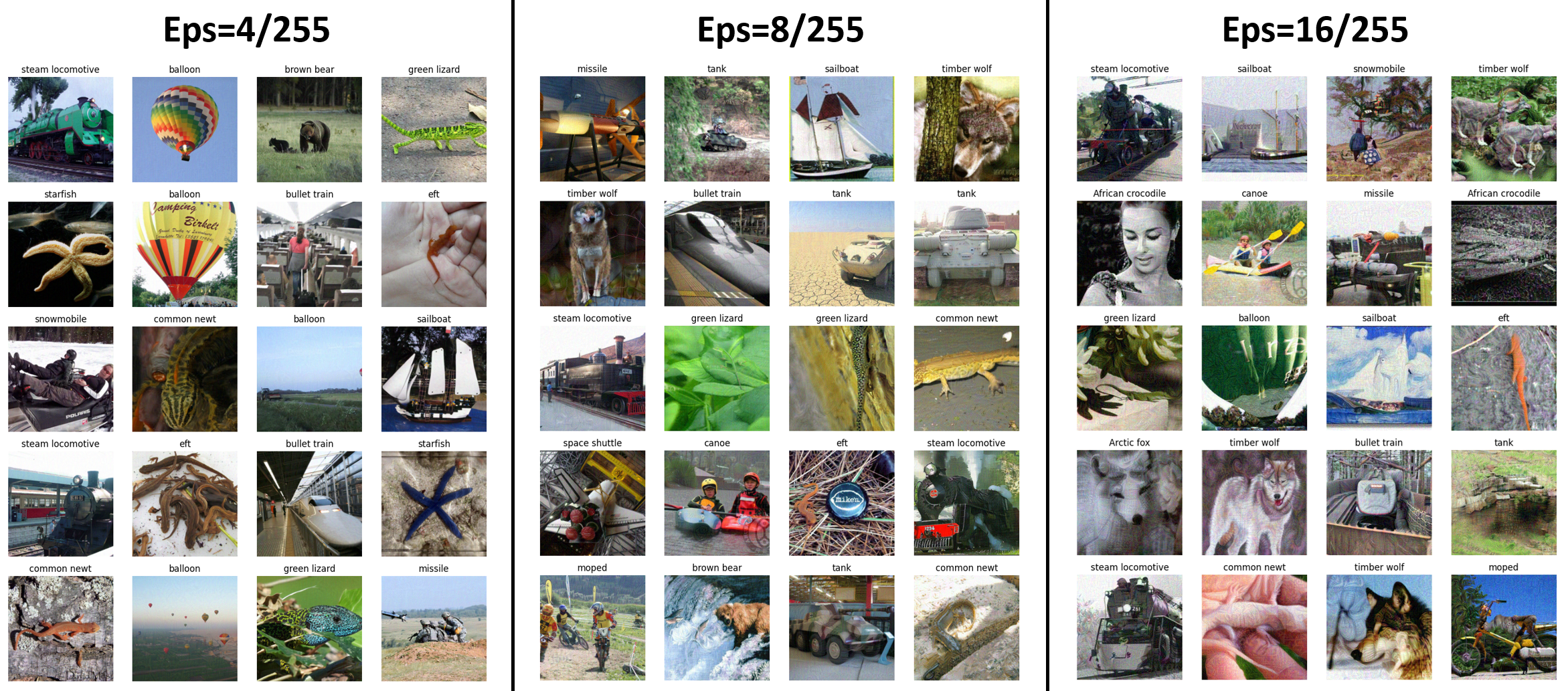}
    \caption{Adversarial examples at different $\ell_\infty$ perturbation epsilons (budgets).}
    \label{fig:adv_examples}
\end{figure}

\begin{figure}[h]
    \centering
    \includegraphics[width=1.35\textwidth, angle=90,origin=c]{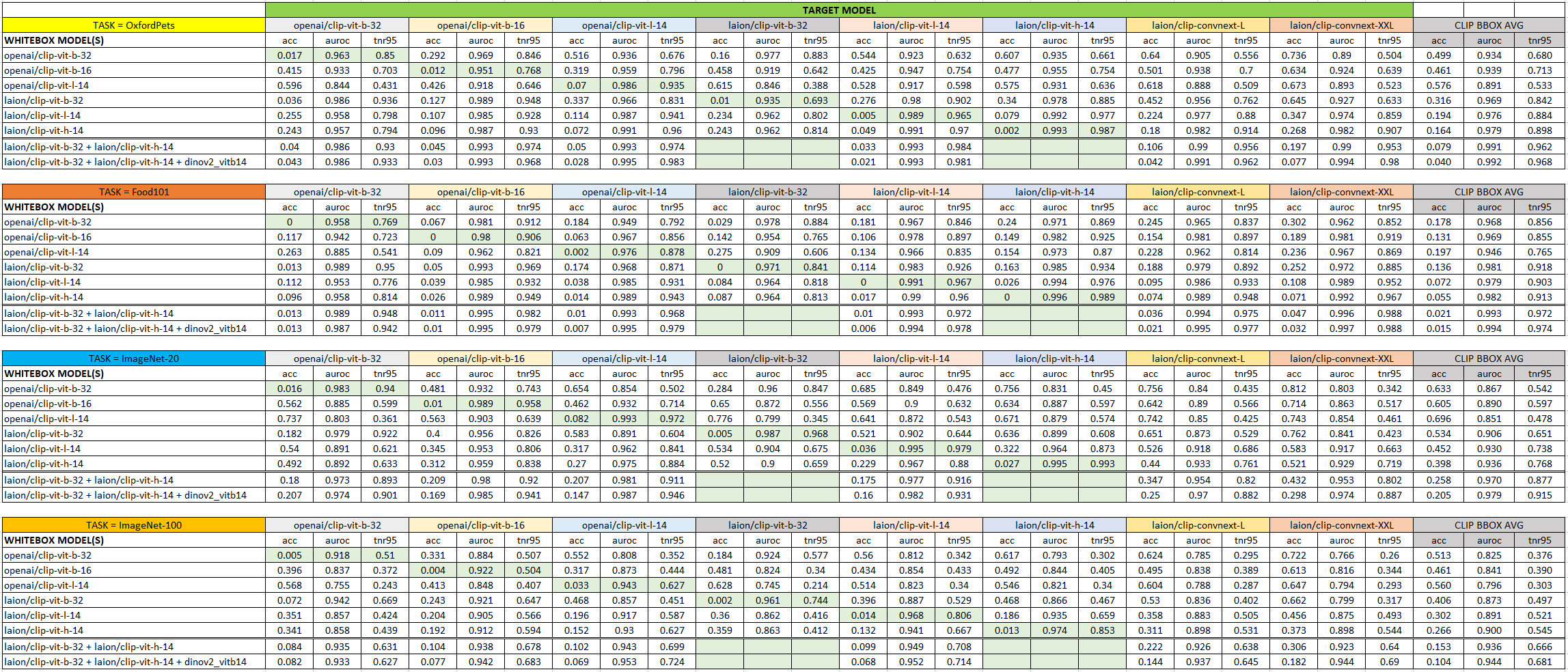}
    \caption{Itemized ID$\rightarrow$OOD attack results.}
    \label{fig:itemized_idood}
\end{figure}

\end{document}